%% file: main.tex
\documentclass{article}

\usepackage[preprint]{neurips_2022}

\usepackage{mathtools}
\DeclarePairedDelimiter{\norm}{\lVert}{\rVert}
\usepackage[utf8]{inputenc} 
\usepackage[T1]{fontenc}    
\usepackage{hyperref}       
\usepackage{url}            
\usepackage{booktabs}       
\usepackage{amsfonts}       
\usepackage{nicefrac}       
\usepackage{microtype}      
\usepackage{xcolor}         
\usepackage{caption}
\usepackage{subcaption}

\title{The Dimpled Manifold Model of Adversarial Examples in Machine Learning}

\author{
  Adi Shamir \\
  Faculty of Math\&CS\\
  Weizmann Institute of Science\\
  Israel \\
  \texttt{adi.shamir@weizmann.ac.il} \\
     \And
   Odelia Melamed \\
   Faculty of Math\&CS \\
   Weizmann Institute of Science \\
   Israel \\
   \texttt{odelia.melamed@weizmann.ac.il} \\
    \And
   Oriel BenShmuel \\
   Faculty of Math\&CS \\
   Weizmann Institute of Science \\
   Israel \\
   \texttt{oriel.benshmuel@weizmann.ac.il}
}

\begin{document}

\maketitle

    


\input{0abstract}

\input{1introduction}
\input{1.5relatedwork}
\input{2mental_image}
\input{3create_dimples}

\input{4mysteries}
\input{06clinganddimple}

\input{08onandoffmanifold}
\input{9summery}
\newpage


\bibliography{bib}

\newpage
\appendix

\input{apdix0-othermodels} 
\input{apdix1-natural_images} 
\input{apdix3.5-lowdimension} 
\input{apdix4-approx_manifold} 
\input{apdix4.5-perpen_exprmnt} 
\input{apdix5-onoffPGD} 
\input{apdix6-NNdetails} 
\input{apdix8-inf_norm}  
\input{apdix2-random_projection} 

\end{document}

%% file: 0abstract.tex
\begin{abstract}
The extreme fragility of deep neural networks, when presented with tiny perturbations in their inputs, was independently discovered by several research groups in 2013. However, despite enormous effort, these adversarial examples remained a counterintuitive phenomenon with no simple testable explanation. In this paper, we introduce a new conceptual framework for how the decision boundary between classes evolves during training, which we call the {\em Dimpled Manifold Model}. In particular, we demonstrate that training is divided into two distinct phases. The first phase is a (typically fast) clinging process in which the initially randomly oriented decision boundary gets very close to the low dimensional image manifold, which contains all the training examples. Next, there is a (typically slow) dimpling phase which creates shallow bulges in the decision boundary that move it to the correct side of the training examples. This framework provides a simple explanation for why adversarial examples exist, why their perturbations have such tiny norms, and why they look like random noise rather than like the target class. This explanation is also used to show that a network that was adversarially trained with incorrectly labeled images might still correctly classify most test images, and to show that the main effect of adversarial training is just to deepen the generated dimples in the decision boundary. Finally, we discuss and demonstrate the very different properties of on-manifold and off-manifold adversarial perturbations. We describe the results of numerous experiments which strongly support this new model, using both low dimensional synthetic datasets and high dimensional natural datasets.
 
\end{abstract}

%% file: 1introduction.tex
\section{Introduction}
\label{sec:introduction}

In 2013 \citet{szegedy2013intriguing} and \citet{biggio2013evasion} independently demonstrated the surprising fact that even the best trained deep neural networks were extremely fragile when presented with tiny adversarial perturbations. This discovery naturally attracted a lot of interest, and many attempts to explain this phenomenon have been proposed over the last nine years: that DNNs are too nonlinear, that they are too linear, that they were trained with an insufficient number of training examples, that adversarial examples are just the rare cases where DNNs err, that images contain robust and nonrobust features, etc. However, none of these qualitative ideas seems to provide a simple, intuitive explanation that can be experimentally tested for adversarial examples' existence and bizarre properties.

\begin{figure}[t]
	\vskip -0.0in
	\begin{center}
		\centerline{\includegraphics[width=\columnwidth]{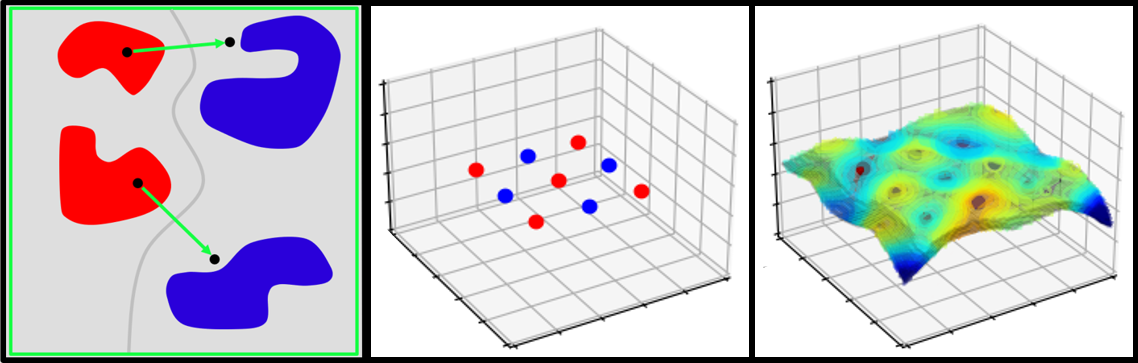}}
		\caption{The simplistic mental image which is often used to describe adversarial examples is shown on the left; in the center, we show a synthetic 2D image manifold with a horizontally arranged chessboard pattern of red and blue training examples in a 3D input space; on the right, we show the actual decision boundary we got by training a DNN on this dataset, described as a topographic map.}
		\label{fig:oldnewmodel}
	\end{center}
	\vskip -0.2in
\end{figure}

This paper aims not to propose new adversarial attacks or defenses but to propose a new comprehensive framework for thinking about adversarial examples. Numerous papers and talks about this subject use some variant of the highly misleading 2D image on the left side of Fig. \ref{fig:oldnewmodel}. In this mental image, the square $[0,1] \times [0,1]$ contains multiple clusters of training images from two classes (denoted by red and blue, respectively). The training aims to create a 1D curved decision boundary (denoted by the grey line) that splits the input space into two (not necessarily connected) parts. Its goal is to place each training example on the correct side of the decision boundary and as far as possible from it in order to maximize the confidence level in its provided label, subject to the limited expressive power of the given DNN. In this mental image, adversarial examples are created by moving the given images along with the green arrows towards some kind of centroid of the nearest training images with the opposite label (as stated, for example, by Ian Goodfellow in his lecture \citet{goodfellowtalk} at time 1:11:54).



For simplicity, we consider in this paper only 2-class classifiers for images that are represented as points in the $n$-dimensional cube $[0,1]^n$, and use $L_2$ norms. This input space is split into two complementary (not necessarily connected) $n$-dimensional regions by the curved $n-1$ dimensional decision boundary.

%% file: 1.5relatedwork.tex
\section{Related work}
While searching for explanations for adversarial examples, several previous papers had already pointed out that the underlying image manifolds have lower dimensions, and that the existence of nearby adversarial examples implies that decision boundaries must pass very close to the given images. For example, \cite{tanay2016boundary} analyzed this question mathematically but only in the linear case; \cite{khoury2018geometry} used geometric arguments to show that adversarial examples leave the manifold, and noted the strong influence of the large number of off-manifold dimensions on the existence of adversarial examples; \cite{stutz2019disentangling} also show these examples are off-manifold, and claimed that on-manifold adversarial examples are generalization errors. All the aforementioned papers presented experiments on synthetic data as well as on small datasets such as  EMNIST, Fashion-MNIST or CelebA. A line of work which used the off-manifold nature of adversarial examples to build new defences and attacks can be found in (\cite{jalal2017robust}, \cite{meng2017magnet}, \cite{samangouei2018defense}). In appendix \ref{appendix:other-models} we discuss other models of adversarial examples and compare them to the one presented in this paper.







%% file: 2mental_image.tex
\section{The new mental image of adversarial examples}
\label{sec:current}

The (well-known) fact which underlies the new conceptual framework is that all the natural images are located on or near some low dimensional manifold (as shown by a huge number of previous papers, from \citet{statnatural} to \citet{pope2021intrinsic}). We can approximate this manifold by using a high quality autoencoding DNN (\cite{bourlard1988auto}, \citet{multilayerAE}) which first compresses the given image into a $k$-dimensional {\em latent space}, and then decompresses this latent space into a $k$-dimensional {\em image manifold} in the $n$-dimensional input space. Note that by using a DNN with ReLU activation functions, we can guarantee that the approximated manifold will be a continuous piecewise linear surface within the image cube, with a well-defined $k$-dimensional basis for the local linear subspace at any given point on it. In typical autoencoders, $k$ can be one to two orders of magnitude smaller than $n$, and their outputs are visually very similar to their inputs. While we do not fully understand the global structure of this manifold, we expect it to be fairly benign. This is because image compression is achieved primarily by eliminating some high spatial frequencies and exploiting the non-uniform distribution of small patches in natural images. 

Next, we note that while decision boundaries are $n-1$ dimensional objects, their quality is judged only by their performance on some natural images within the tiny $k$-dimensional image manifold. In fact, during the training, networks try to utilize the vast perpendicular subspace (on which they are not judged) to make it easier for them to place the decision boundary within the image manifold correctly. We visually demonstrate this mental image for the low dimensional case of $k=2$ and $n=3$ in the middle part of Fig. \ref{fig:oldnewmodel}, in which a 2D image manifold floats in the middle of the 3D cube of possible inputs at height $z=0.5$. According to the simplistic mental image, when we add to the input space a new third dimension we could expect a 1D grey decision boundary as in the left image to be extended upwards and downwards (for all $0 \leq z \leq 1$) into a vertical 2D wall that separates the red and blue clusters. However, in our synthetic simulation of a DNN on the synthetic dataset in the middle figure (and in the vast majority of real image settings), we got the decision boundary depicted on the right side of Fig. \ref{fig:oldnewmodel}, which clings very closely to the whole image manifold, except for shallow dimples that gently undulate below the red clusters, and above the blue clusters. We call this conceptual framework the {\em Dimpled Manifold Model} (DMM) and note that it is based on two testable claims about how decision boundaries evolve during the training process:

\begin{enumerate}

\item{Training a DNN proceeds in two distinct phases: a (typically fast) {\it clinging phase} which brings the decision boundary very close to the image manifold, followed by a (typically slower) {\it dimpling phase} which creates shallow bumps in the decision boundary that try to move the boundary to the correct side of the training examples}
\item{To help the training, DNNs develop large gradients in the confidence levels in the vicinity of the training examples, which point roughly perpendicularly to the image manifold}

\end{enumerate}

To support this model, we will do three things: Explain intuitively why the training process is likely to prefer such a decision boundary over other possibilities, show that it provides simple explanations for the counterintuitive properties of adversarial examples, and finally verify it by direct experimental measurements over multiple synthetic and natural datasets.

%% file: 3create_dimples.tex
\section{Why DNN's are likely to create dimpled manifolds as decision boundaries}

\begin{figure}[ht!]
  \centering

         \includegraphics[scale=0.25]{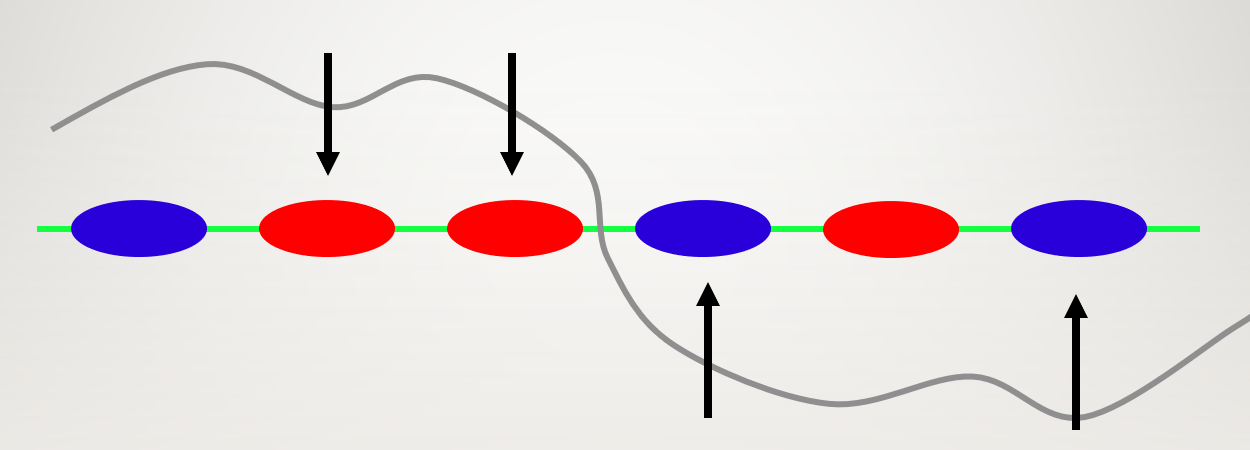}
         \caption{Bending the initial decision boundary to accommodate misclassified training examples applies mostly inward pressure that pushes the boundary closer to the image manifold}
         \label{fig:hammerblows}
\end{figure}

A randomly initialized DNN is likely to create a randomly oriented initial decision boundary. Let us consider once again our new mental image of a flat horizontal 2D image manifold at middle height in a three-dimensional input cube. Let us assume that the initial decision boundary passes well above the image manifold in a certain region of this manifold (which contains training images of both classes). In the first epoch, we try to push the decision boundary ``downwards'' in the gradient's direction (which is roughly perpendicular to the current decision boundary) at any red cluster. At the same time, we do nothing at blue clusters (since they are already on the correct side of the boundary). Similarly, in regions where the initial decision boundary passes well below the image manifold, we will want to leave the decision boundary unchanged at red clusters and move it ''upwards'' at blue clusters (see Figure \ref{fig:hammerblows}).

Thinking about the decision boundary as a thin sheet of pliable metal, we bend it over multiple epochs by applying tiny hammer blows to regions we want to move up or down. However, most of these blows are directed off the image manifold. Eventually, we expect this process to yield a metal sheet that conforms to the shape of the image manifold, except in the vicinity of red training examples (where we create downward-pointing dimples) or blue training clusters (where we create upward-pointing dimples). 

Note that by developing a large derivative in the vertical direction, the network can bend the decision boundary more gently. This bending makes it easier to gain accuracy with a simpler decision boundary that uses shallower dimples that pass on the correct sides of neighboring training examples of opposite classes. In addition, the gentle bending of the decision boundary can create larger bumps that cover multiple training examples along with regions between these examples (where no hammer blows are applied), leading to a possible generalization phenomenon. Finally, the above description explains why different architectures, complexities, and classifiers are likely to have decision boundaries that almost coincide with the image manifold (but can be quite different on other parts of the input domain, in which there are no training or test images). 

%% file: 4mysteries.tex
\section{Explaining the counterintuitive properties of adversarial examples}
The biggest difference between the two mental images is how we think about adversarial examples. In the old mental image, adversarial examples were created by going horizontally towards the nearest training examples with the opposite label, which are all very far away. In the new mental image, adversarial examples are created by going vertically a tiny distance $\epsilon$ towards the dimpled decision boundary and then continuing another $\epsilon$ on the other side. This model provides the following simple explanations:

\begin{enumerate}

\item
What are these adversarial examples? How can it be that next to any cat image, there is also an image of guacamole and vice versa? The answer is that all the real cat and guacamole images reside on the tiny image manifold. However, ``below'' and ``above'' the manifold, there are vast half-spaces of pseudo-images recognized by the network as cats and guacamoles even though they do not look like ones. The adversarial examples we generate are such pseudo-images. Note that when we consider multi-class classifiers in an $n$ dimensional input space, there are multiple $n-1$ dimensional decision boundaries between pairs of classes. In this case, any two decision boundaries have roughly perpendicular ``up'' and ``down'' directions.

\item Why are the adversarial examples so close to the original images? As explained above, DNNs prefer to have large perpendicular derivatives in order to have shallower dimples that make it easier to undulate the decision boundary around the training examples gently. The tiny distance is a direct consequence of this large gradient since it suffices to move a short distance to significantly affect the confidence level.

\item
Why don't the adversarial perturbations resemble the target class? When we use an adversarial attack to modify a cat into guacamole, why doesn't the perturbation we use look green and mushy? Most adversarial perturbations look like a featureless small-magnitude random noise. In the new mental image, we are moving roughly perpendicularly to the direction of guacamole images. For example, if a unit vector towards the nearest guacamole image is $(1,0,0, \ldots, 0)$, then a random unit vector in a perpendicular direction has the form $(0,x_2,x_3, \ldots , x_n)$, in which each $x_i$ is a tiny positive or negative value around $O(1/ \sqrt{n})$. Such an adversarial perturbation looks (especially in $L_\infty$ norm) like the random salt and pepper perturbations we see in the standard demonstrations of adversarial examples, rather than a depiction of guacamole.



\item
It has been experimentally demonstrated that more robust networks tend to be less accurate. Why do robustness and accuracy trade-off? In the new model, ease of training and the existence of nearby adversarial examples are two sides of the same coin. When we train a network, we keep the images stationary and move the decision boundary around them by creating dimples; when we create adversarial examples, we keep the decision boundary stationary and move the images to its other side. Allowing a large perpendicular derivative makes the training easier since we do not have to bend the decision boundary around the training examples sharply. However, such a large derivative also creates very close adversarial examples. Any attempt to robustify a network by limiting all its directional derivatives will make it harder to train and thus less accurate.

\end{enumerate}

\subsection{Adversarial training just deepens the dimples}
Adversarial training is a training method and a well-researched area that was proven many times to improve the robustness of NN (starting from \citet{szegedy2013intriguing} and thousands of others since then). In each epoch, we create adversarial examples using the training data and a specified attack and train the network over these examples as train data. 

The new conceptual model enables us to analyze the effect of adversarial training of DNNs from a new perspective. We take each training image during an adversarial training and create a nearby adversarial example. Let us assume that all these adversarial examples were successfully generated and that they are all shifted the same distance $2 \epsilon$ in the vertical direction with respect to the original image ($\epsilon$ to reach the decision boundary, and another $\epsilon$ to get high confidence in the opposite decision). By adding the adversarial examples (with visually correct labels) to the original training set, we create a new image manifold whose thickness had been increased to $2 \epsilon$.
When we adversarially train a new network with only the adversarial counterparts, its dimples will have to go an extra distance of about $2 \epsilon$ deeper than before. This easily explains why it is harder to adversarially train a network (since we have to use more hammer blows to push the dimples further away from the image manifold), and why we may lose accuracy for regular test images (since the decision boundary has to make sharper turns at the deeper dimples and may miss some small clusters).

\begin{figure}
  \centering
  \includegraphics[width=\textwidth]{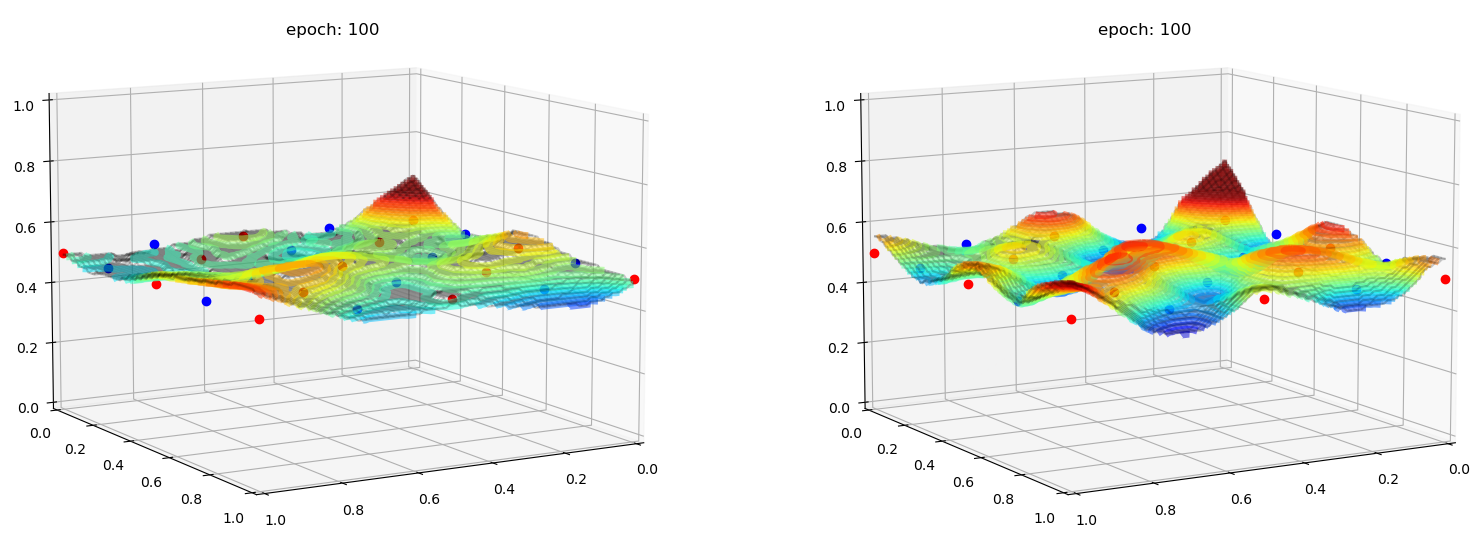}
  \caption{A 3D binary decision boundary of two NNs with the same architecture and same random initialization. The NN in the left image was trained with "clean" 2D data; the right NN was adversarially trained using a single-step $L_2$ PGD attack. }
  \label{3dadvtrain}

\end{figure}

We look at the clinging phase as a phase induced by the data manifold and the dimpling phase as induced by the specific labels of the training data. We now explain the adversarial training process using the DDM model and the two phases above. As adversarial examples are within a short distance from the data points (and therefore from the data manifold), one can assume that the clinging phase will not change much due to the definition of adversarial examples as the train set. 
Therefore, the major effect of the adversarial training is within the dimples phase. After the clinging phase, the adversarial direction is perpendicular to the data manifold. Therefore, the effect of the adversarial training on the decision boundary is a deepening of the dimples, as one can see in Figure \ref{3dadvtrain}. Note that when the dimples get deep enough, the best adversarial direction (one calculated using a gradient with respect to the input) changes. While the shortest way to cross the boundary was previously almost orthogonal to the manifold, the dimples are deeper after adversarial training. As a result, the gradient will have a more significant on-manifold component. However, taking a slightly larger step in the same off-manifold direction will also result in an adversarial example, just a bit further. 

This conclusion helps us explain why creating adversarial examples for robust networks will result in partly on-manifold examples that have some semblance to the adversarial class, as shown previously in \cite{tsipras2018robustness}.

\subsection{Adversarial training with incorrect labels}
A particularly interesting experiment was performed in 2019 by \citet{madrysexpr} from Madry's group at MIT (simplified here into a 2-class setting):

\begin{enumerate}

\item
Start with a training set of images $s_1, \ldots s_m$ of cats and guacamoles, and train a deep neural network $N_1$ to recognize them well.

\item

Use $N_1$ to create from each training example $s_i$ an adversarial example $t_i$ of the opposite class.

\item

Train a fresh deep neural network $N_2$ (with a possibly different architecture/size and new random initialization) using only the adversarially modified images $t_1, \ldots ,t_m$, along with the {\it visually wrong} (ie, target) class labels.

\end{enumerate}

The new network $N_2$ never saw anything that visually resembles a cat being labeled a cat: all the slightly modified cats were labeled as guacamole, and all the slightly modified guacamoles were labeled as cats.
However, when the new network $N_2$ is applied to the original test images, it has good accuracy, labeling a significant fraction of the cat images as cats.

\begin{figure}[t]
	\vskip -0.0in
	\begin{center}
		\centerline{\includegraphics[scale=0.4]{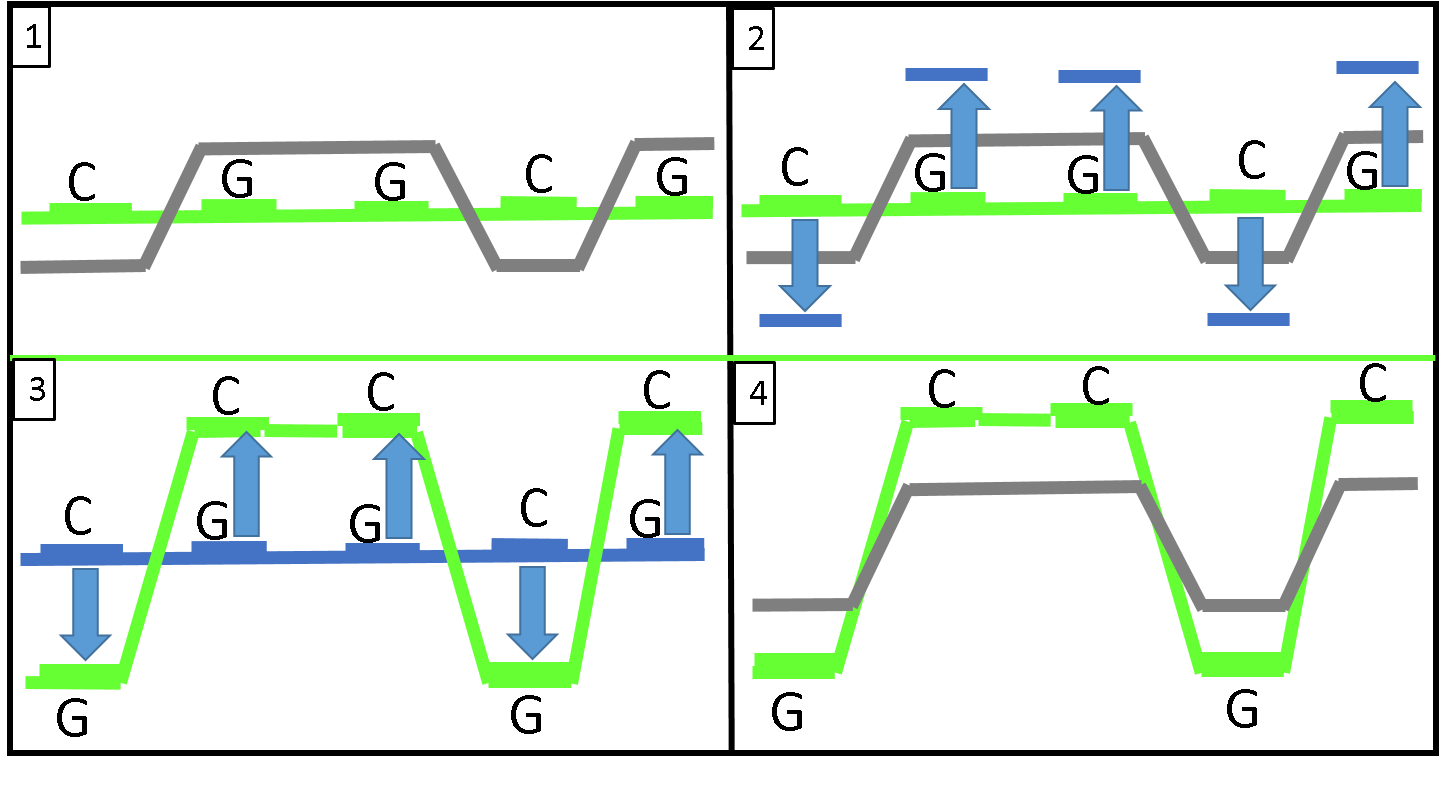}}
		\caption{A schematic side view of the green original image manifold, the green distorted image manifold, and the grey decision boundaries}
		\label{fig:Madry_all}
	\end{center}
	\vskip -0.2in
\end{figure}

The original paper tried to explain this highly surprising result by distinguishing between robust and non-robust features in any given image, claiming that they behave differently under adversarial training. However, it is unclear what makes some particular features more robust than others and how significant these differences are. Our alternative explanation (which does not necessarily contradict their explanation) is summarized in Fig. \ref{fig:Madry_all} which describes a series of 2D vertical cuts through the input space. Basically, we claim that the original training of $N_1$ created a decision boundary which is close to the manifold but shifted up and down a distance of $\epsilon$ away in the direction of the labels. On the other hand, the training of $N2$ consisted of clinging the decision boundary to an image manifold which had been shifted away a distance of $2 \epsilon$ in the same directions and then bending it back a distance of $\epsilon$ in the opposite directions during the dimpling phase, which creates very similar decision boundaries in $N_1$ and $N_2$. This is a simpler geometrical interpretation of the experiment, compared to their qualitative explanation of the existence of robust and non-robust features.

In Fig. \ref{fig:Madry_all}, we show the cat'ness decision boundary created when $N_1$ is trained on the original training examples in part 1 of the figure; it is a dimpled line that passes a distance $\epsilon$ under all the cats and the same distance $\epsilon$ above the guacamoles.

As depicted in part 2 of Figure \ref{fig:Madry_all}, when $N_1$ is used to create adversarial examples, all the images generated from cats are moved down a distance $2 \epsilon$ to make $N_1$ think that they are guacamoles with a high level of confidence. All the images generated from guacamoles are moved up a distance of $2 \epsilon$ to make $N_1$ think that they are cats with a high confidence level.

When we use only the adversarial examples as training images for $N_2$, they form a warped new image manifold which is depicted by the new green line in part 3 of Fig. \ref{fig:Madry_all}. Note that we consistently mislabel all the adversarial examples by using their adversarial labels. After training $N_2$, we get the new grey cat'ness decision boundary in part 4 of Fig. \ref{fig:Madry_all}), which passes a distance $\epsilon$ below all the cat-labeled training images, and a distance $\epsilon$ above all the guacamole-labeled training images. By comparing parts 1 and 4 of the figure, it is easy to see that the two cat'ness decision boundaries in them are actually very similar to each other. It is thus not surprising at all that when the strangely trained $N_2$ is applied to the original test images, it has good accuracy in recognizing cats!

%% file: 06clinganddimple.tex
\section{Clinging and Dimpling}
\begin{figure}[ht!]
\centering
\begin{subfigure}[b]{\linewidth}
\centering
\includegraphics[width=\textwidth]{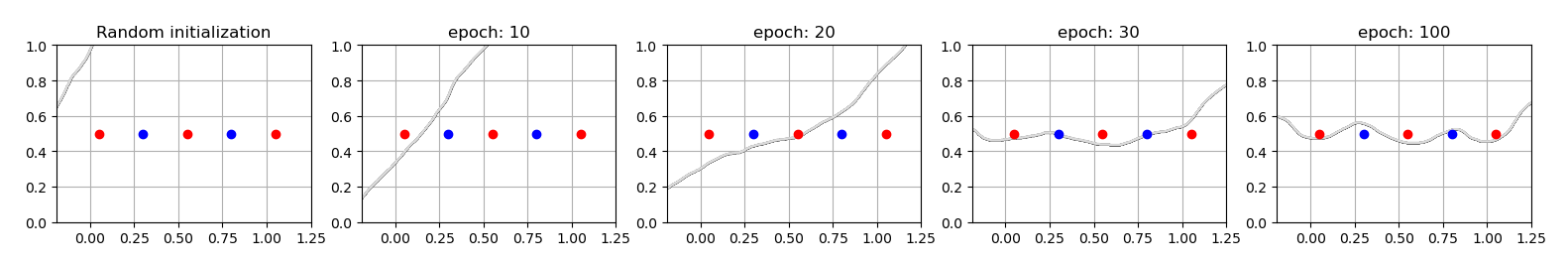}
\caption{A 1D binary decision boundary evolving in a 2D input space, first clinging to the 1D data manifold then dimpling to fit the data. The red and blue colors are the labels of the two classes. The grey lines at the specified epochs show the evolution of the decision boundary after the network is randomly initialized. 
}
\label{fig:1ddimples}
\end{subfigure}
\begin{subfigure}[b]{\linewidth}
\centering
\includegraphics[width=\textwidth]{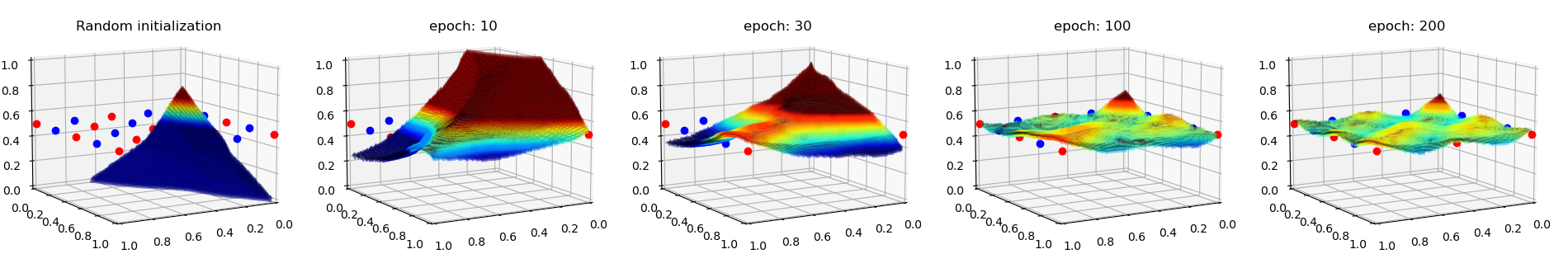}
\caption{A 3D binary decision boundary evolving in a 2D input space, first clinging to the 2D data manifold then dimpling to fit the data. The red and blue colors are the labels of the two classes. The decision boundary is colored by its $z$ values, making the dimples easily visible.}
\label{fig:2ddimples}
\end{subfigure}
     
     \caption{The evolution of decision boundaries during training epochs in real training sessions.}
\end{figure}

In this section we demonstrate the Dimpled Manifold Model in a low-dimensional synthetic setting, in which the manifold is precisely defined, and the clinging and dimpling phases of the training can be directly visualized. We train a simple 2-layers network over two low dimensional datasets: (1) a 1D diagonal line in a 2D input space (Figure \ref{fig:1ddimples}) and (2) a 2D linear subspace in a 3D input space (Figure \ref{fig:2ddimples}). In both training processes, we can see that the decision boundary is getting very close to the manifold in the first few epochs, and later a gentle dimpling (using the extra dimensions) process fits the given labels. 

The main difficulty in demonstrating the correctness of the Dimpled Manifold Model for high-dimensional datasets and deep neural networks is that it is not possible to directly visualize what the decision boundary looks like in a very high dimensional space. However, a simple test of this model is to measure the average distance between the test examples and the evolving decision boundary during the various training epochs. Due to the high dimensions, a random initialization is likely to be much closer to the data manifold. Thus, the initial random boundary is much closer to the test images than in the lower dimensional cases.

\begin{figure}[ht!]
\centering
\centering
\includegraphics[scale=0.35]{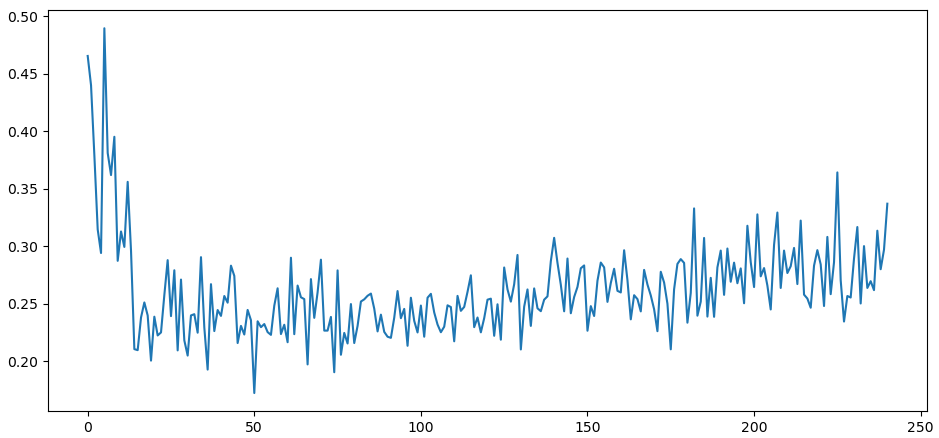}
\caption{Distance to the decision boundary during training: The x-axis represents the epoch number, and the y-axis represents the average distance between the boundary and the test-set images
}
\label{fig:cifartrain}
\end{figure}

Such an experiment with the CIFAR10 dataset and ResNet50 10-classes classifier is described in Figure \ref{fig:cifartrain}, where the distance graph shows a two-phase behavior in which this distance rapidly decreases during the first few epochs as the decision boundary clings to the image manifold, and then fluctuates over many epochs without further decreases, as the dimples grow in various directions. Note that in these high dimensional settings, the initial distance to the boundary is already quite small.

This characterization of the training process is further supported by an information-theoretic analysis described in \citet{tishbi}. In that paper (and its accompanying video), the authors demonstrated what happens when they train a DNN on a small synthetic dataset. During training, there was a first phase in which the main effect was to increase the mutual information between the neurons in the DNN and the training inputs, and then a very different second phase in which the main effect was to increase the mutual information between the neurons in the DNN and the desired labels. In our framework, this directly corresponds to a first phase in which the decision boundary increases its knowledge of the image manifold by clinging to it, and a second phase in which it increases its knowledge of the labels of the training examples by creating appropriate dimples.

%% file: 08onandoffmanifold.tex
\section{Adversarial examples off and on the manifold}
\label{section:onandoff}
We conducted experiments on the MNIST, CIFAR10 and ImageNet datasets (\citet{lecun2010mnist}, \citet{CIFAR10}, \citet{IMAGENET}). For each data set, we approximated the low-dimensional natural-image manifold $M$ using an autoencoder with a high compression rate. We computed its local linearization around each test-set image $x$ (see Appendix \ref{appendix:approx-manifold} for further details and dimension specifications). This approximation may, in fact, overestimate $k$, yet it demonstrates well the dimpled manifold model.\\

\paragraph{Perpendicularity to the image manifold}
The perpendicularity of adversarial examples to the image manifold has been investigated geometrically and theoretically before, and it had even inspired several defense mechanisms. As one can see in the last epoch's decision boundary in Figures \ref{fig:1ddimples} and \ref{fig:2ddimples}, our low dimensional experiments demonstrate the same perpendicularity  (experiments with MNIST, CIFAR10, and IMAGENET datasets which demonstrate this perpendicularity for real datasets are described in appendix \ref{appendix-perpendicularity}). Using this property and our locally approximated image manifold, we can analyze the difference between on-manifold and off-manifold adversarial perturbations.\\

\begin{figure}[ht!]
  \centering
         \includegraphics[width=\textwidth]{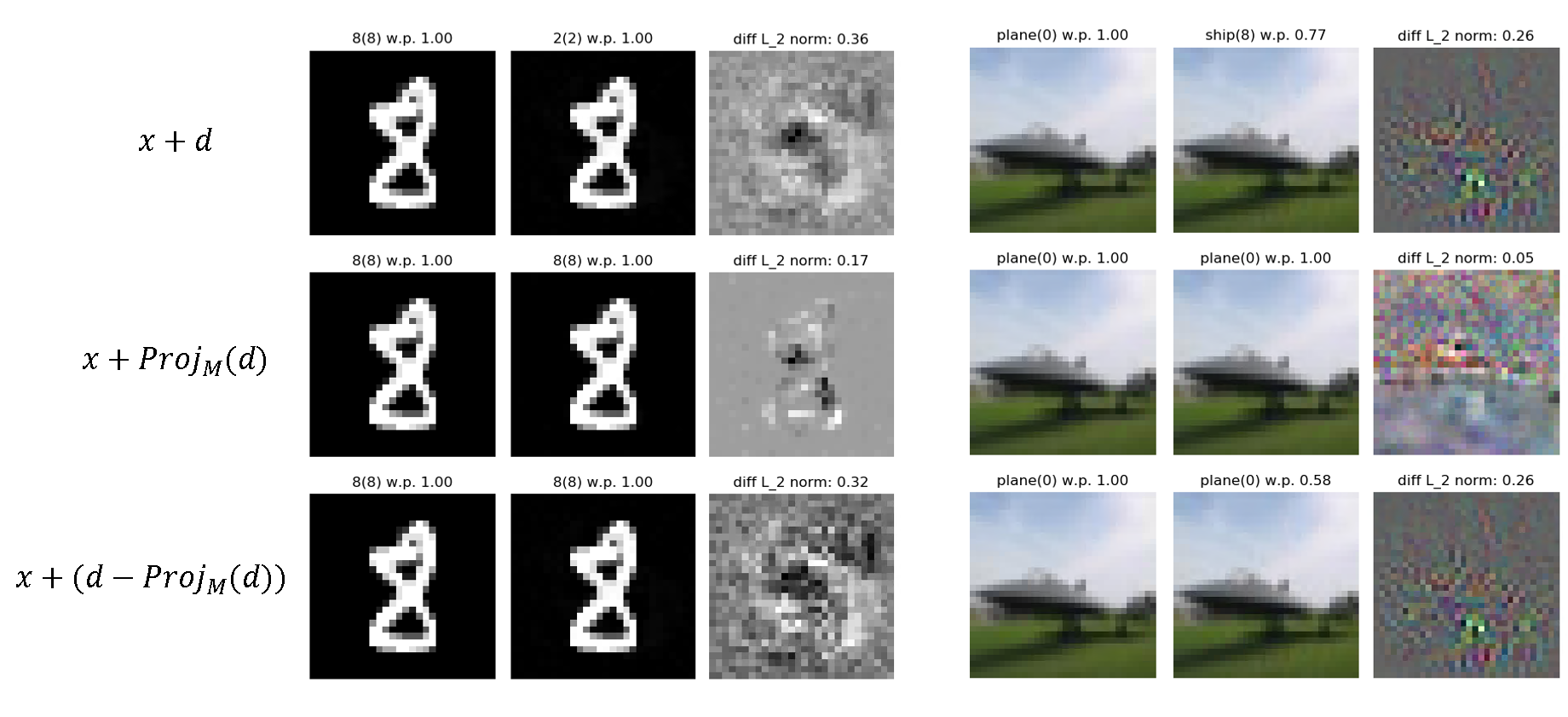}
       \caption{Left: MNIST - changing 8 to 2. Right: CIFAR10 - changing plane to ship}
    \label{fig:mnist-proj-m}

\end{figure}

We first consider adversarial examples crafted on MNIST and CIFAR10 classifiers using PGD attack (\citet{PGD}). For an image $x$, we started with calculating an adversarial example $x+d$ using a multi-step PGD attack. We projected $d$ onto the locally approximated linear manifold $M$, denoting this projection $Proj_{M}(d)$. Therefore, the resultant off manifold projection is $(d - Proj_{M}(d))$. In Figure \ref{fig:mnist-proj-m}, for a test-set image $x$ we show the original adversarial example $x+d$ (first row), the projected on-manifold example $x+Proj_{M}(d)$ (second row), and the projected off-manifold example $x + (d - Proj_{M}(d))$ (last row). The columns in the figure from left to right are: the natural image $x$, the perturbed image, and the perturbation itself (maximally amplifying its entries to the full range of $[0,1]$ to make it visually clearer).

In the left part of Figure \ref{fig:mnist-proj-m}, we can see that while the large off-manifold perturbation is hard to interpret, the small on-manifold perturbation is doing what the human visual system would expect, trying to open the two loops in the ``8''  to get a ``2'' by blackening the relevant white segments in them. In the right Figure \ref{fig:mnist-proj-m} we show the PGD attack and its projections on a CIFAR10 test-set image, demonstrating similar behavior. Here, the on-manifold projection tries to change the plane visually into a ship - painting the grass in blue to resemble the sea. Additional examples from these three datasets can be found in appendix \ref{appendix:natural-image-exam}.



\subsection{On-manifold adversarial examples reveal features from the target class}
\cite{tsipras2018robustness} separates adversarial examples created with small epsilon and clean-trained network and one created using large epsilons and robust networks. While the former looks like noise, the latter creates examples that resemble the target class. Our experiment uses on-manifold and off-manifold adversarial examples to demonstrate a very similar phenomenon, indicates a possible connection between an on-manifold gradient and robustness.

Starting from an image $x$, we used the PGD attack with an extra constraint: before projecting each step onto the epsilon-step sphere, we projected it on or off the approximated local linear manifold $M$ (see Appendix \ref{appendix:onoffattacks} for attack details). In this way, we generated on and off-manifold adversarial examples. Off-manifold adversarial examples are significantly closer to the test image than the on-manifold examples and only slightly further than non-constrained examples. More details about adversarial distances can be found in appendix \ref{appendix-perpendicularity}.

\begin{figure}[ht!]
  \centering

         \includegraphics[width=0.7\textwidth]{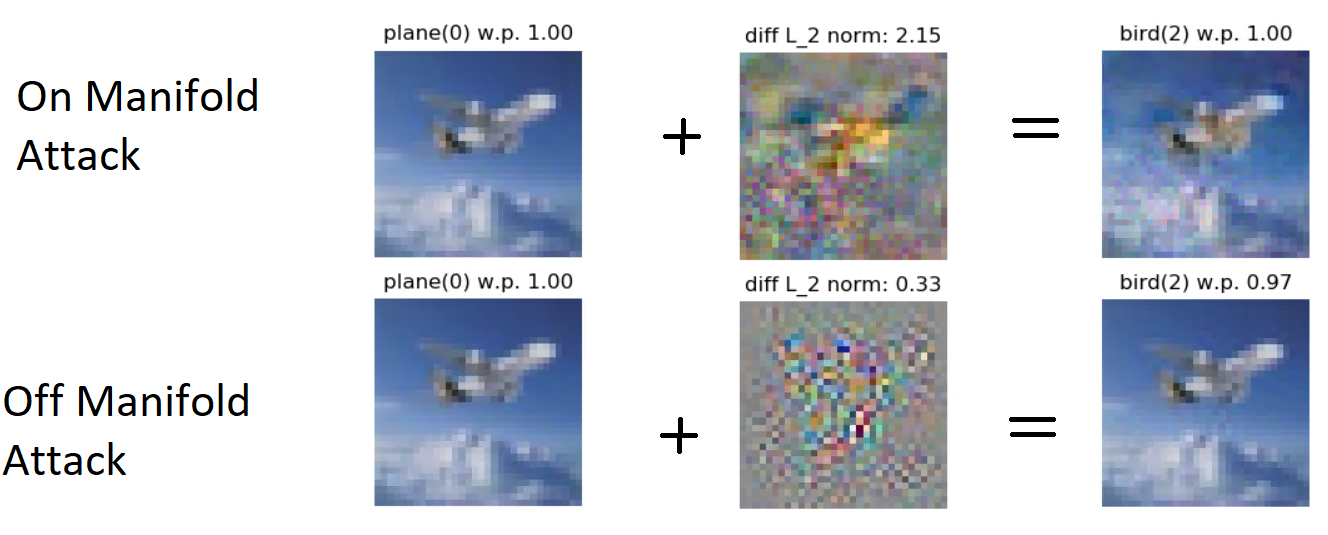}

       \caption{CIFAR10 - on and off manifold attacks changing a plane into a bird}
    \label{fig:mnist-fig}

\end{figure}

The inherent difference between the on and off manifold examples in these settings is very clear. In Figure \ref{fig:mnist-fig}, the on-manifold adversarial perturbations are visually meaningful. In contrast, the off-manifold perturbations consist of low-level noise surrounding the object's general area: At the top of Figure \ref{fig:mnist-fig} one can see that the network is creating a bird by coloring the plane and shortening its wings. Additional examples for this experiment can be found in appendix \ref{appendix:natural-image-exam}.
    
%% file: 9summery.tex
\section{Open problems}

In this paper we demonstrated that the new DMM conceptual framework can provide simple and testable explanations for most of the counterintuitive properties of adversarial examples. The main remaining open problem is why adversarial perturbations are often transferable among classifiers of different types, structures, sizes, initializations, and training examples. We hope that the common shape of the image manifold in all these cases will help clarify this issue as well.




%% file: apdix0-othermodels.tex
\section{Appendix- Comparison to other suggested models}
\label{appendix:other-models}
Our Dimpled Manifold Model describes the existence and properties of adversarial examples from a new geometric perspective. The best known alternative perspective is the one proposed in \cite{ilyas2019adversarial}, which assumes the existence of two types of features in any given image. The robust features are the human-recognizable features of the image, while the non-robust features are non-indicative features that humans ignore or never notice but which can be used by the classifier. The two perspectives can be viewed as complementary rather than contradictory, since they try to describe the same phenomenon using different languages: we use geometric intuition whereas \cite{ilyas2019adversarial} takes a more human-centric approach.

Clearly, adversarial perturbations which are mostly orthogonal to the image manifold lead to off-manifold adversarial examples. We can thus associate the off-manifold dimensions to the "non-robust features" of \cite{ilyas2019adversarial}. On the other hand, the on-manifold dimensions correspond more closely to the human-noticeable "robust features". In section \ref{section:onandoff} and Appendix \ref{appendix:natural-image-exam} we demonstrate an on-manifold perturbation which changes human interpretable features of the natural images in a targeted attack from the source class to the target class. This indicates that the manifold dimensions identified by our auto-encoder (which might be a subset or a superset of the local dimensions of the real image manifold) are more closely associated with human-recognizable features of the image. 

A similar effect is also shown in \cite{tsipras2018robustness} in a robustness context. They show that when they take an epsilon step for a very large epsilon using the robust network's gradient, the input image seems to develop an appearance which is more related to the target class. The dimpled manifold model can easily explain this effect. During adversarial training, we claim that the dimples get deeper. Therefore the off-manifold distance to the decision boundary gets larger, while the on-manifold distance to the boundary does not change much. This can change the gradient's direction since relatively speaking it has a more significant on-manifold component. 

Note that the Dimpled Manifold Model and the robust vs. non-robust features model explain adversarial training differently. The robust vs. non-robust features model claims that the adversarial training makes the network learn something essentially different - in clean training, it tends to learn the non-robust features, while in adversarial training it pays attention mainly to the robust features. The difference is intuitively related to the fact that the networks notice that the non-robust features don't influence the labeling. According to our model, the decision boundary learned is similar in the two learning strategies. The only difference is that in the adversarial training strategy, the dimples become deeper, a property that has a significant effect on the local gradient at each data point but a pretty light impact on the high dimensional decision boundary.

The model of boundary tilting presented in \cite{tanay2016boundary} describes the decision boundary in simple settings of only two classes in 2 data blobs. If zoomed in to the area between two specific data points, our two-dimensional experiment demonstrates experimentally the tilting described in the paper.

The high dimensional geometric property related to the concentration of measure described in \cite{mahloujifar2019curse}, \cite{gilmer2018adversarial} and others, where the decision boundary is likely to be within a short distance to the data manifold is highly related to our model. We show a clinging phase in a low-dimensional setting in which the decision boundary is getting closer to the data manifold. This high dimensional clinging effect is demonstrated in our clinging example for the training process of the CIFAR10 dataset. In this experiment, one can see that the initial distance to the nearest point on the boundary is indeed quite close. 

The comparison of robustness to adversarial examples and random noise pointed in \cite{fawzi2016robustness} and others can also be rephrased from our model's point of view. As most input dimensions are off-manifold dimensions, random noise in the input space is also mainly orthogonal to the manifold. Therefore, training using a random noise will also deepen the dimples in off-manifold dimensions, which can be compared to the similar effect caused by adversarial training, and may explain the robustness for adversarial examples. 

%% file: apdix1-natural_images.tex
\section{Appendix - additional natural image examples}
\label{appendix:natural-image-exam}
In this section we provide additional examples from our natural image experiments.

\subsection{Projected adversarial vector visual examples}

In the following figures, we show the on-manifold and off-manifold projections of adversarial perturbations $d$ created by a PGD attack for typical test-set images $x$. for a test-set image $x$ we show the original adversarial example $x+d$ (first row), the projected on-manifold example $x+Proj_{M}(d)$ (second row), and the projected off-manifold example $x + (d - Proj_{M}(d))$ (last row). The columns in the figure from left to right are: the natural image $x$, the perturbed image, the perturbation itself (maximally amplifying its entries to the full range of $[0,1]$ to make it visually clearer), and visualization figure of the change in the logits. In each one of these logit change figures, the $i$-th colored line describes the evolution of the $i$-th logit $f_i$ as we move along a straight line between the original image $x_0$ and the adversarial example $x_1$ generated from $x_0$ (i.e., the vertical axis is the value of logits $f_i((1-t)x_0+tx_1)$, and the horizontal axis is the parameter $t$ that moves from 0 to 1, in 100 equally spaced intervals)
 
In all these examples, the image created by the off-manifold projection $x+(d- Proj_M(d))$ is almost adversarial (i.e., it almost flips the class associated with the max logit), while the on-manifold projection $x+ Proj_M(d)$ barely changes the output logits (i.e., it has almost no adversarial effect). In addition, in all the examples, one can see that the on-manifold projection of the adversarial perturbation is visually meaningful, trying to modify $x$ into an image with the characteristic features of the target class.

\begin{figure}[ht!]
  \centering
  \includegraphics[width=\textwidth]{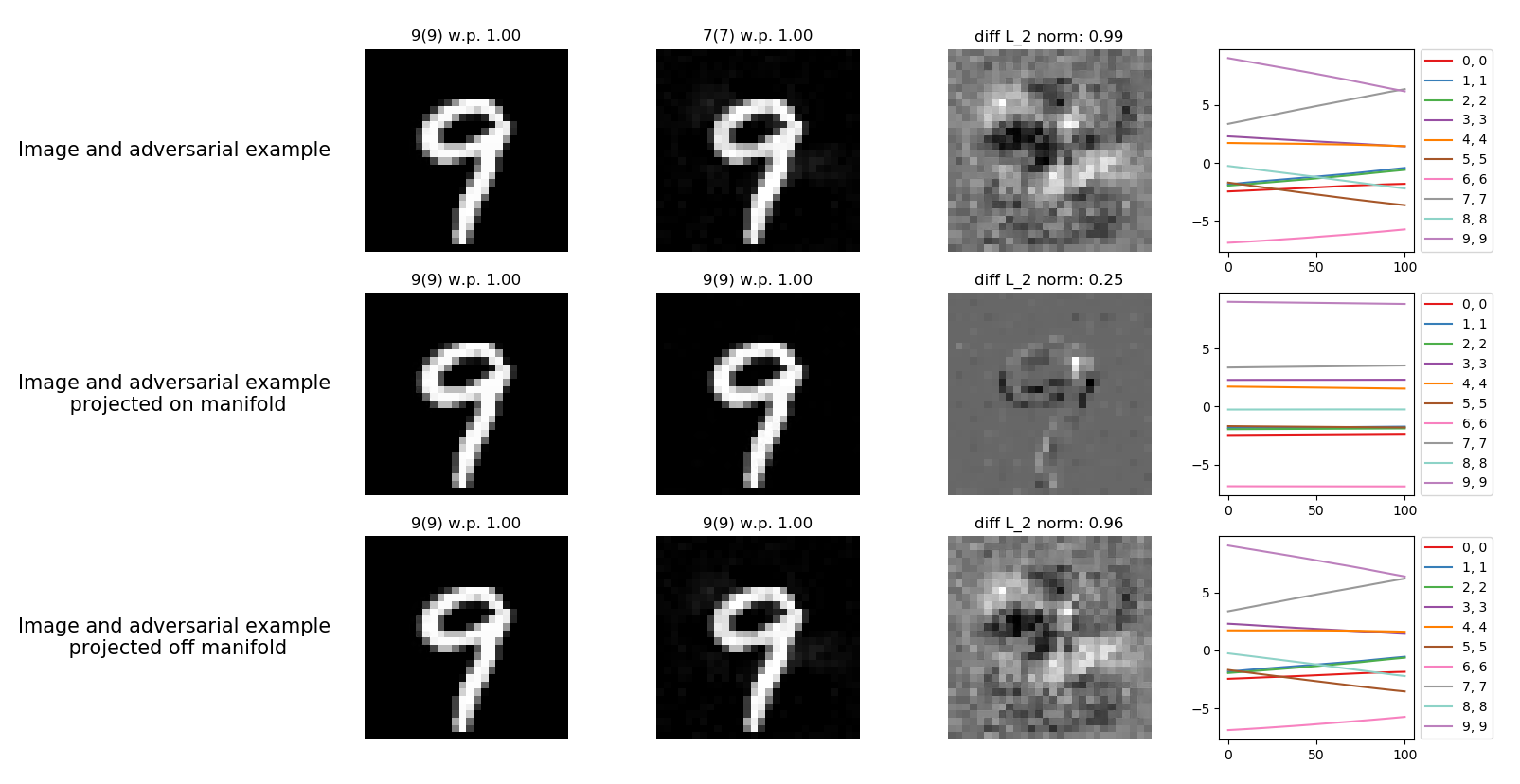}
  \caption{MNIST - The figure 9 changed into 7. The on manifold projection is trying to delete (blacken) the bottom part of the ring in the 9 in order to create a 7-like figure.}
\end{figure} 

\begin{figure}[ht!]
 \centering
 \includegraphics[width=\textwidth]{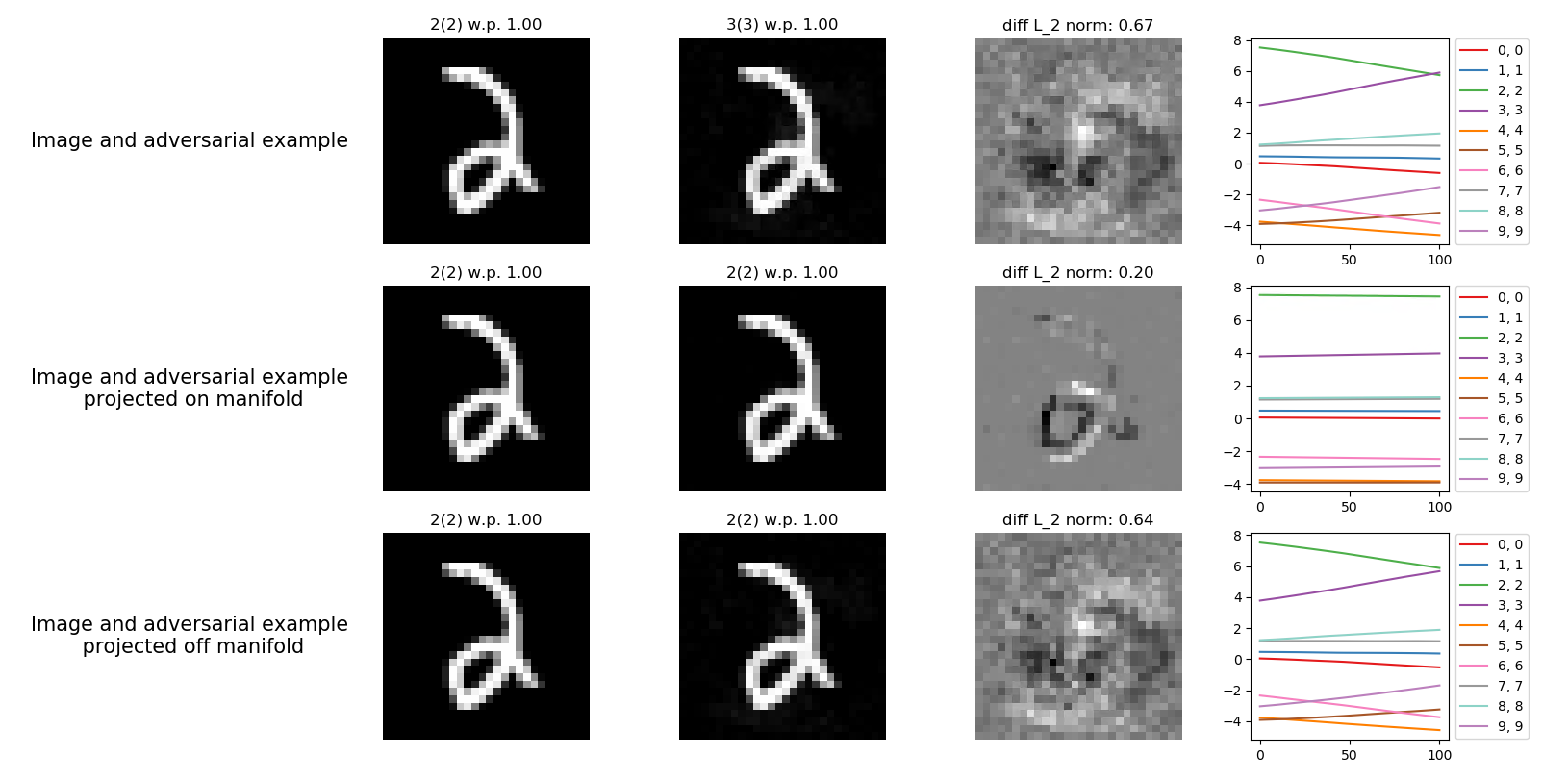}
 \caption{MNIST - The figure 2 changed into 3. The on-manifold projection is trying to delete (blacken) the left part of the circle in the 2 figure in order to open it. It also tries to increase the distance between the two bottom horizontal lines left from the circle by adding a white line outside and a black line inside and tries to eliminate the short protrusion on the right side of the 2.}
\end{figure} 

\begin{figure}[ht]
  \centering
  \includegraphics[width=\textwidth]{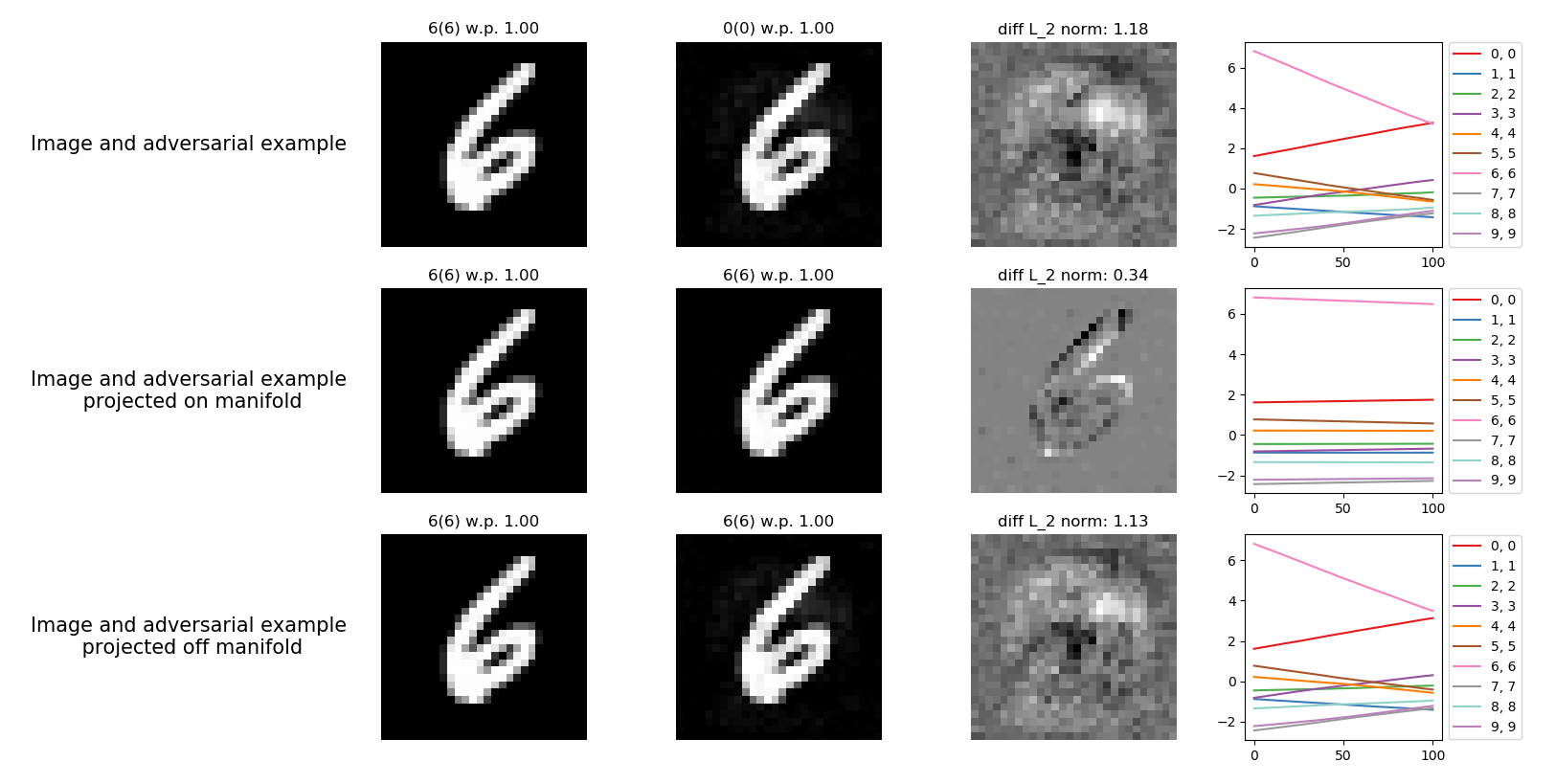}
  \caption{MNIST - The figure 6 changed into 0. The on-manifold projection is trying to close the 6 figure's circle to create a 0 figure.}
\end{figure}
\begin{figure}[ht]
 \centering
 \includegraphics[width=\textwidth]{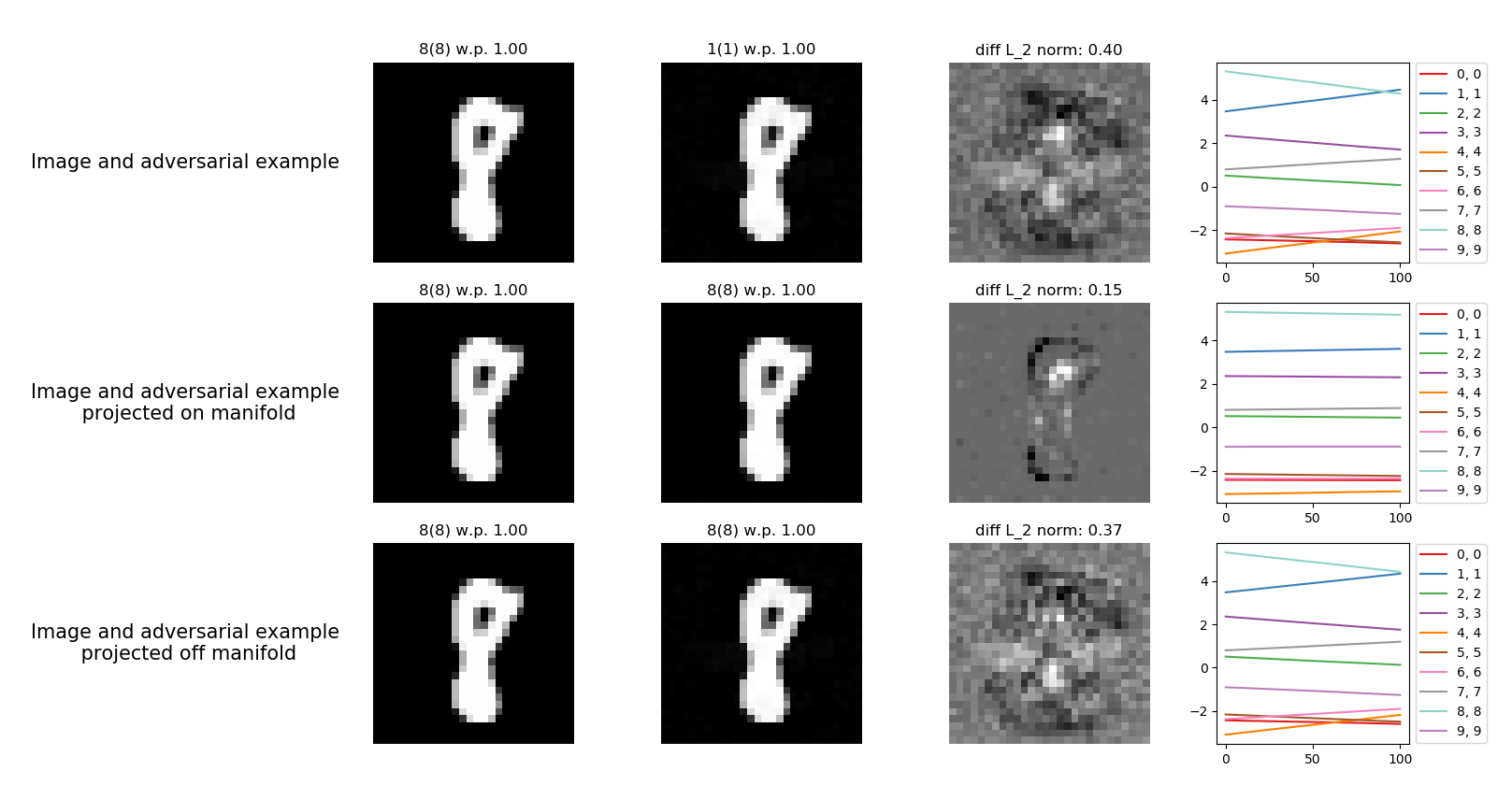}
 \caption{MNIST - The figure 8 changed into 1. The on-manifold projection is trying to fill-in with white the top ring of the 8 figure to create a solid vertical line,  which is the characteristic feature of a 1 figure.}
\end{figure}

\begin{figure}[ht!]
      \centering
     \includegraphics[width=\textwidth]{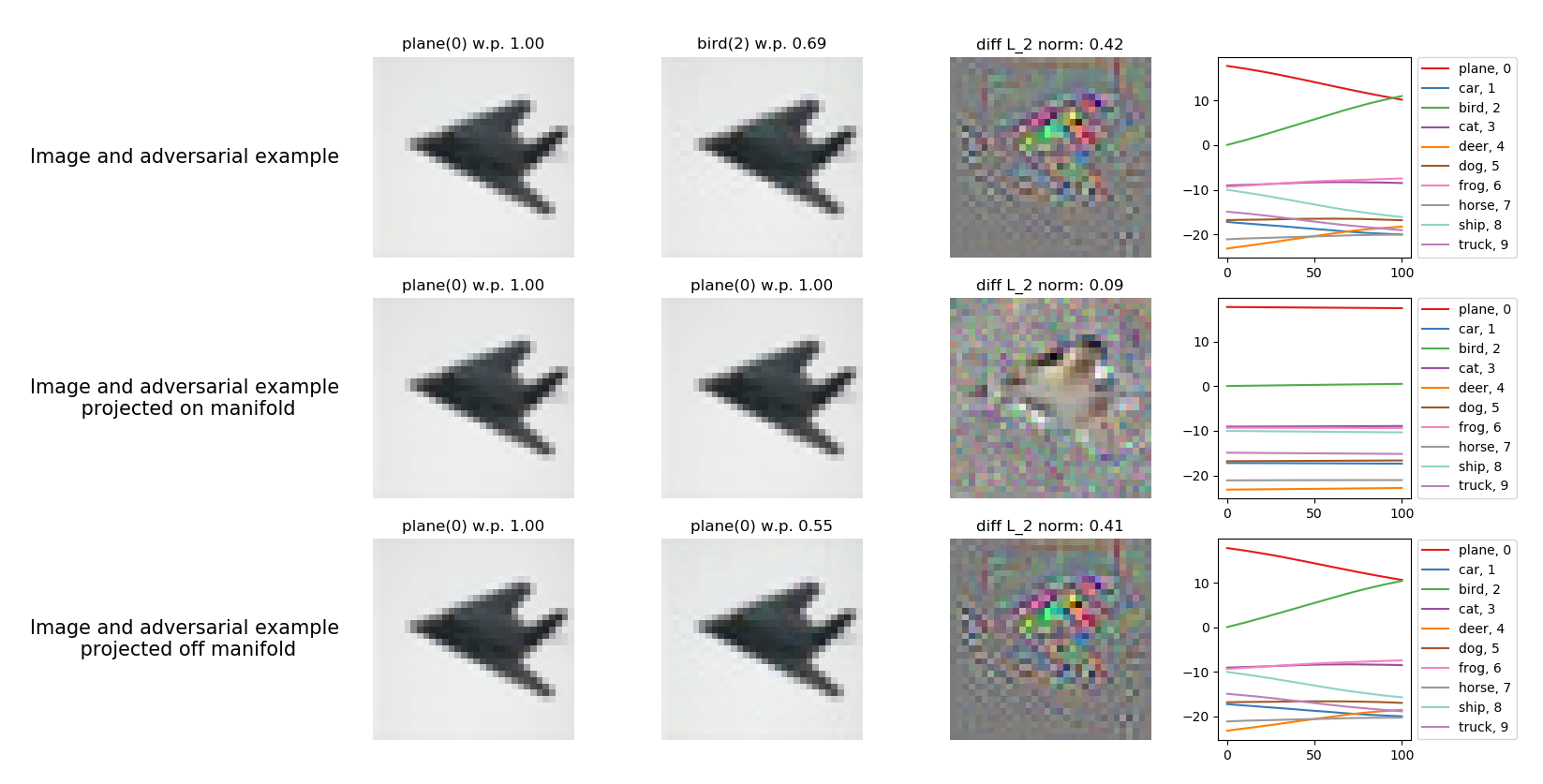}
     \caption{CIFAR10 - a plane is changed into a bird. In the on-manifold projection's row, a bird-shaped and bird-colored perturbation is added to the plane in order to "paint" it like a bird while keeping the overall shape and background.}
     \label{fig:cifarproj1}
\end{figure} 

\begin{figure}[ht!]
     \centering
     \includegraphics[width=\textwidth]{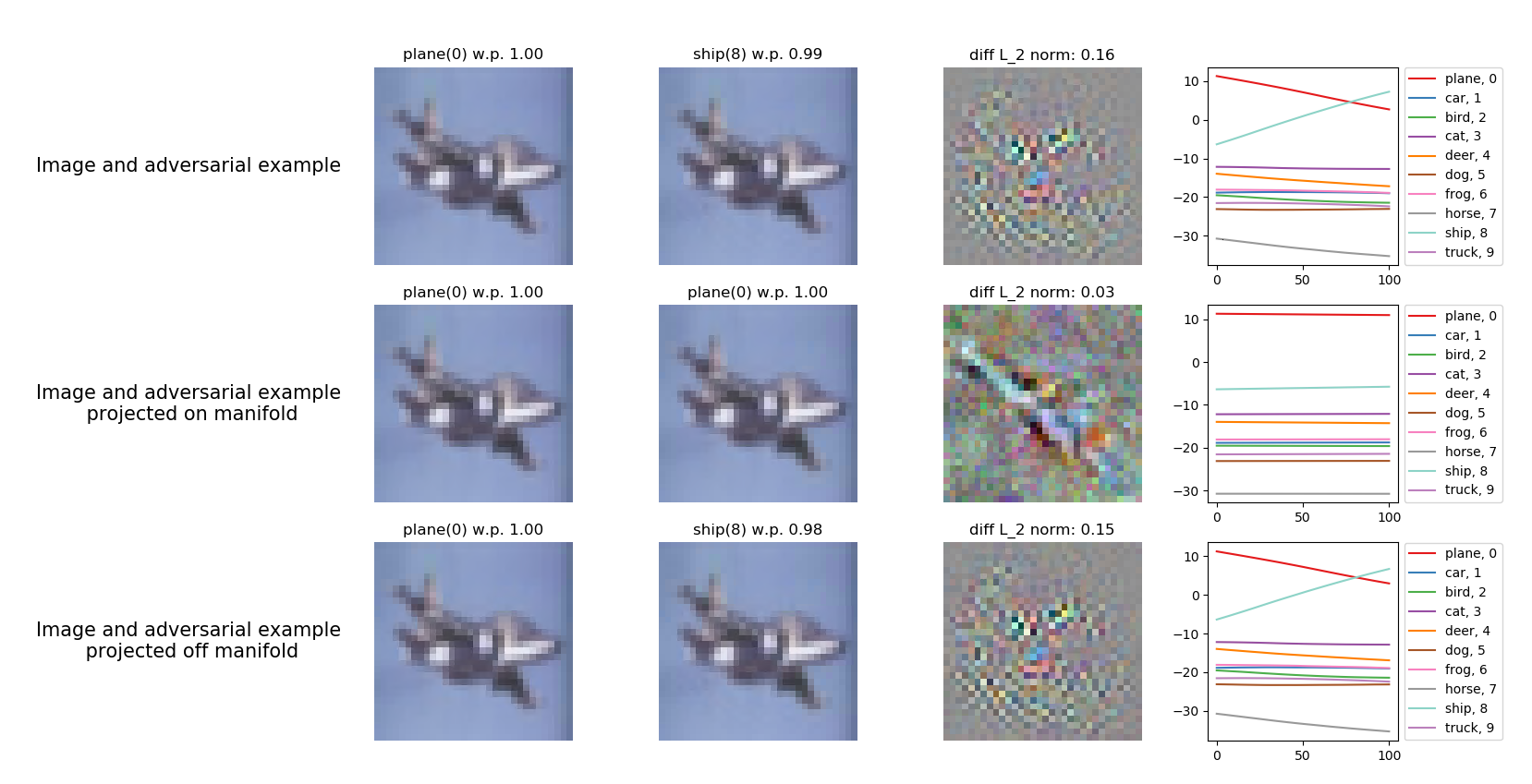}
     \caption{CIFAR10 - a plane is changed into a ship. In the on-manifold projection's row, the wings of the plane are emphasized to turn the plane into a diagonal ship viewed from above.}
\end{figure} 

\begin{figure}[ht!]
     \centering
     \includegraphics[width=\textwidth]{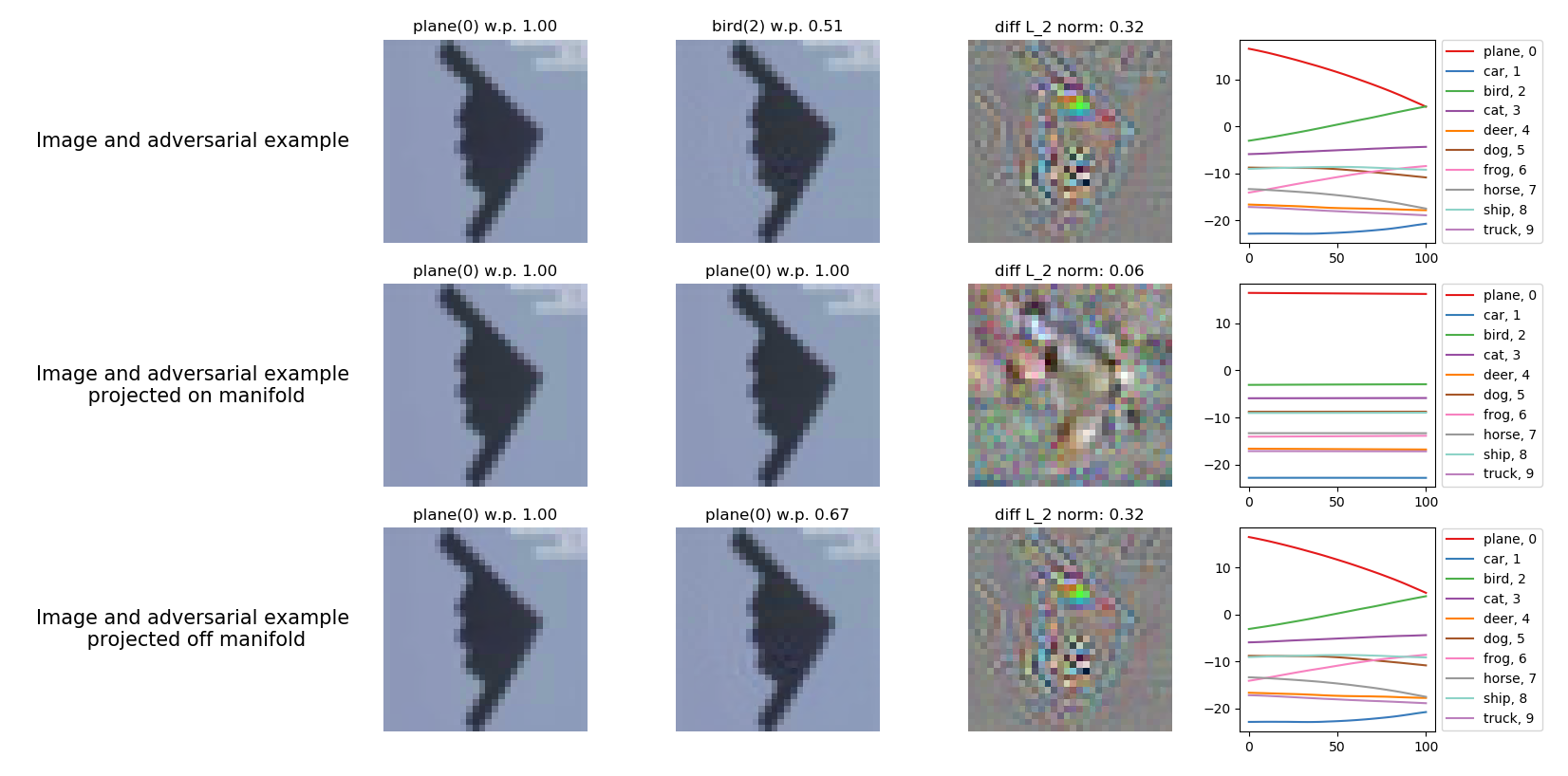}
     \caption{CIFAR10 - a plane is changed into a bird. As in Figure \ref{fig:cifarproj1}, the plane in "painted" like a bird.}
 \end{figure} %
 
 \begin{figure}[ht!]
     \centering
     \includegraphics[width=\textwidth]{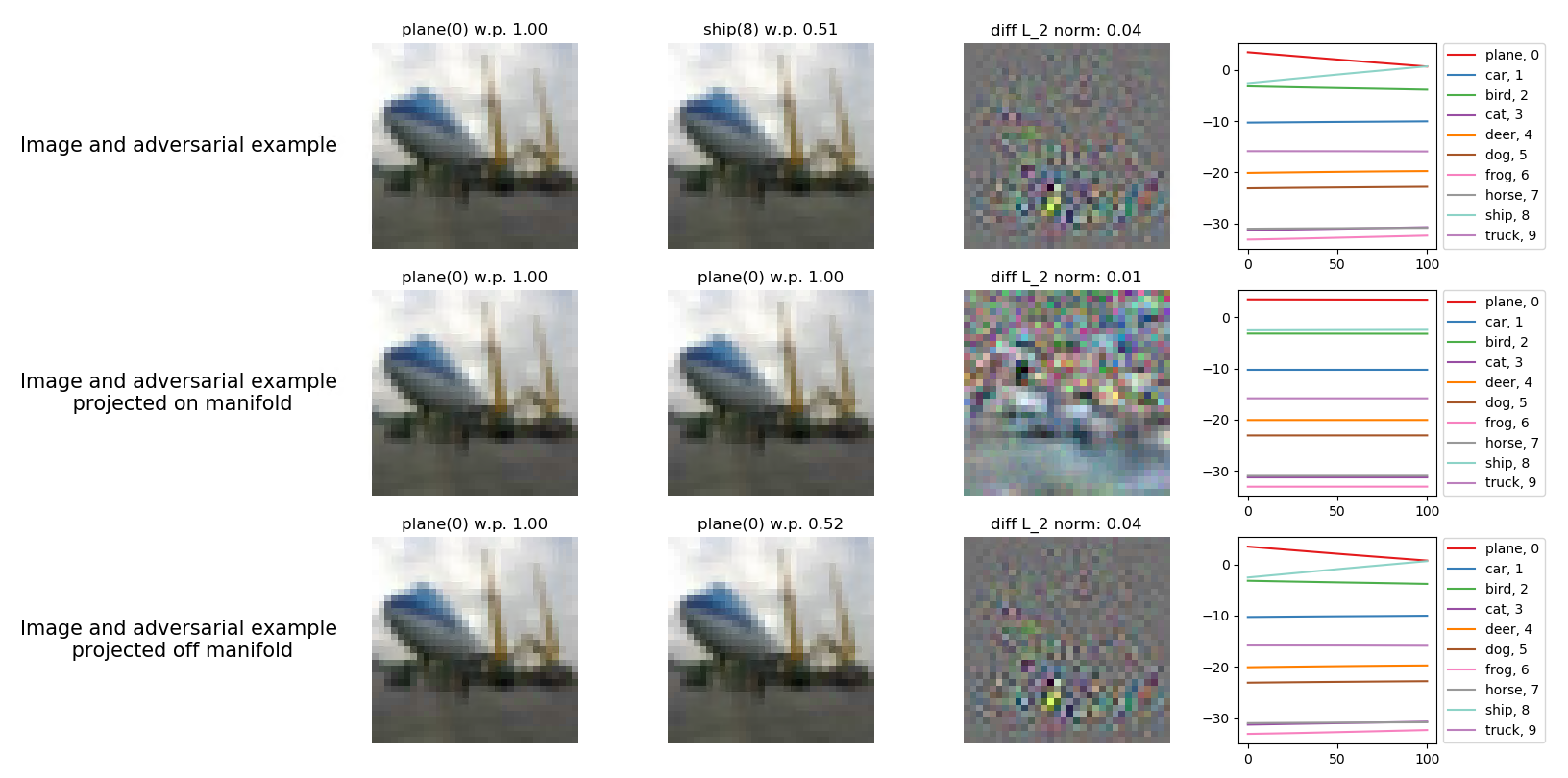}
     \caption{CIFAR10 - a plane is changed into a ship. The on-manifold perturbation is "painting" the plane's runway in blue, to resemble a sea.}
 \end{figure} %
 
\begin{figure}[ht!]
     \centering
     \includegraphics[width=\textwidth]{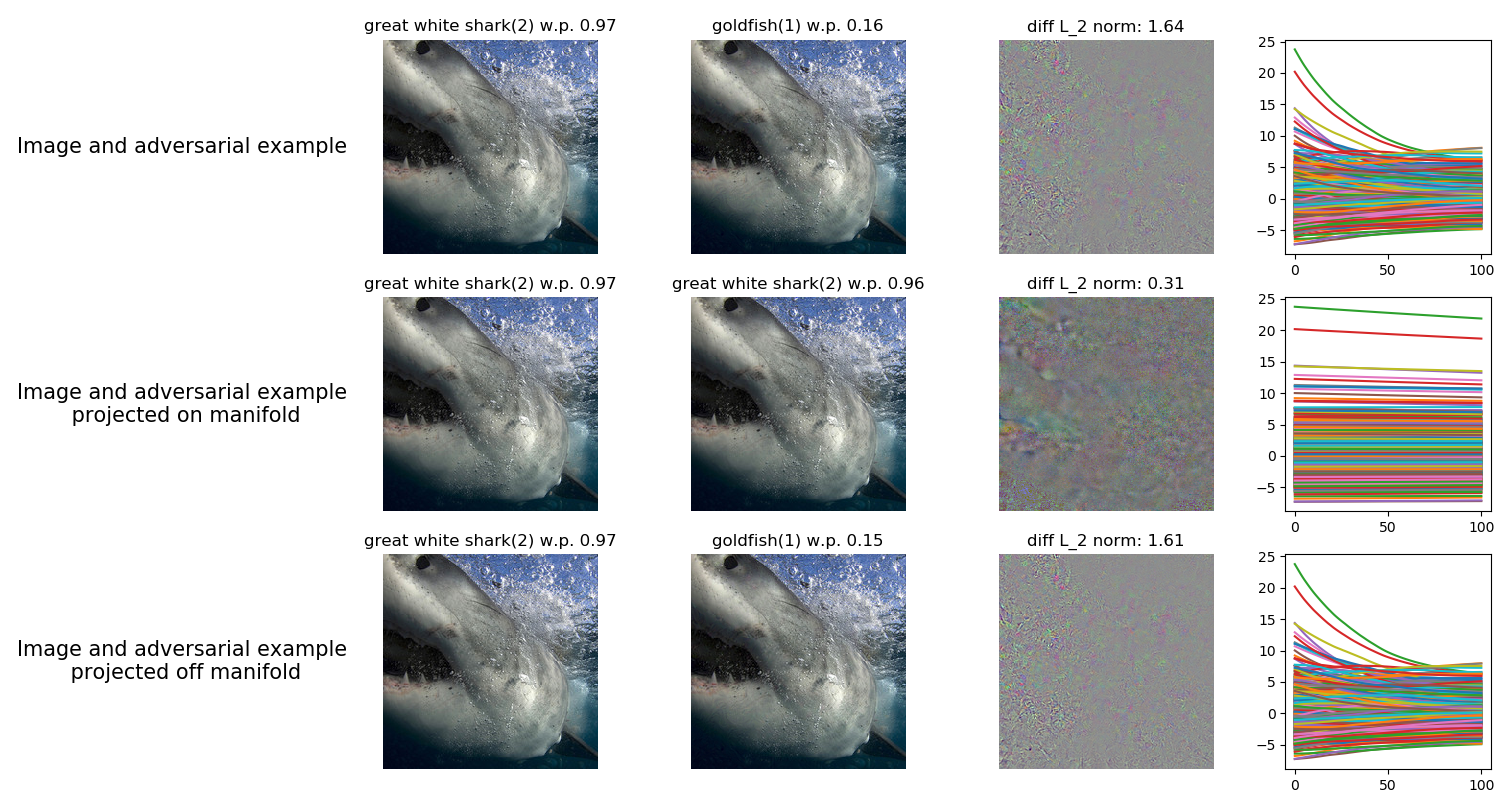}
     \caption{ImageNet - a white shark is changed into a goldfish. A small orange cluster is created on the left. See further explanation in \ref{subs:imagenet}.}
\end{figure} %

\clearpage 

\subsection{Separate on-manifold and off-manifold PGD attack examples}
For each dataset, for a test-set image $x$, we run all three attacks - regular PGD, on-manifold PGD, and off-manifold PGD. We look at the different adversarial perturbations for each attack. We note that the on-manifold attack creates a perturbation that tries to visually change the image to belong to the target class, but its adversarial distance is much larger than for the off-manifold attack.

\begin{figure}[ht!]
  \centering
  \includegraphics[width=\textwidth]{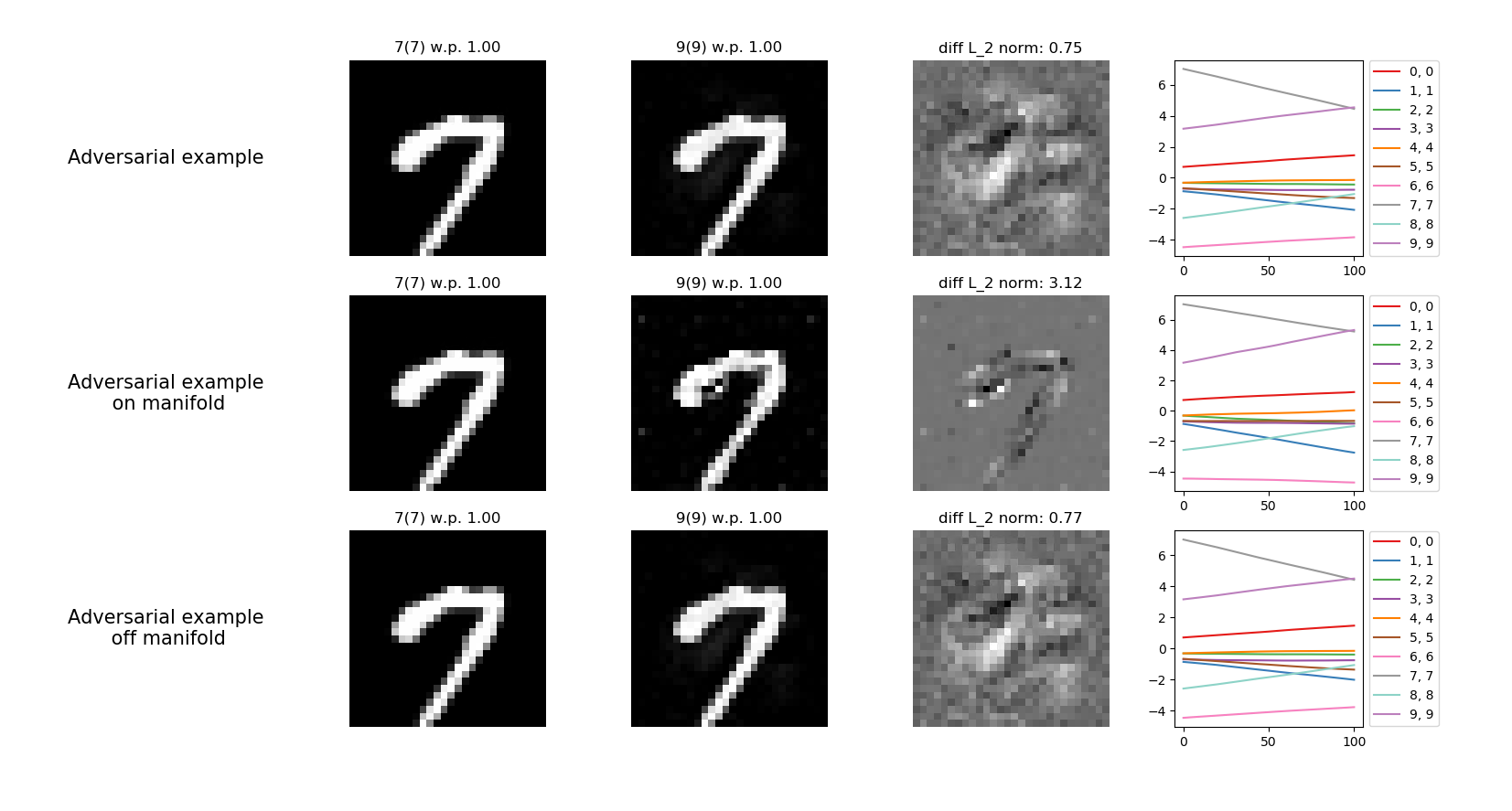}
  \caption{MNIST separate attacks - changing 7 into 9. The on-manifold perturbation is adding a circle at the top of the 7.}
\end{figure}

\begin{figure}[ht!]
  \centering
  \includegraphics[width=\textwidth]{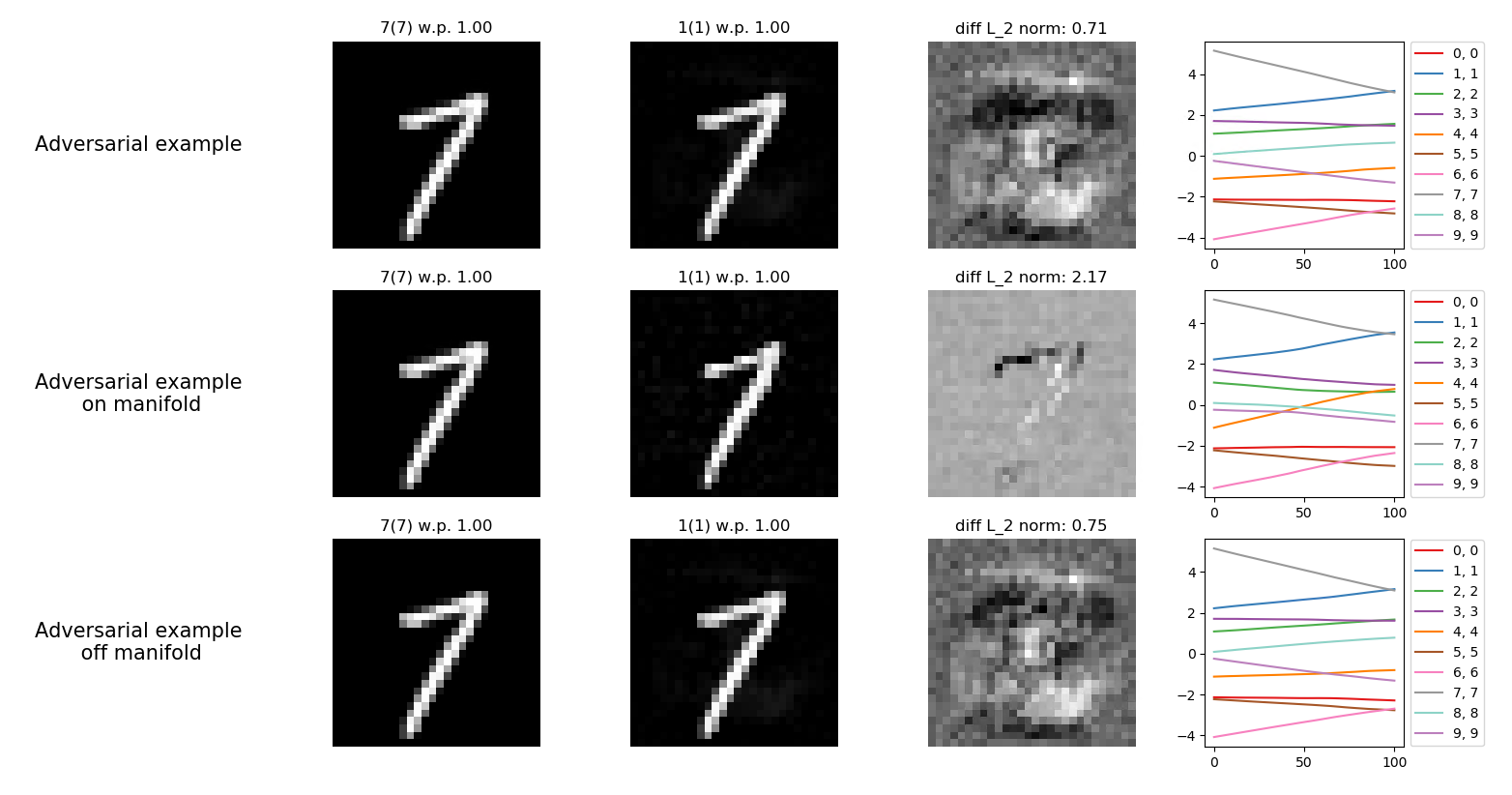}
  \caption{MNIST separate attacks - changing 7 into 1. The on-manifold perturbation is shortening the upper horizontal line.}
\end{figure}

\begin{figure}[ht!]
  \centering
  \includegraphics[width=\textwidth]{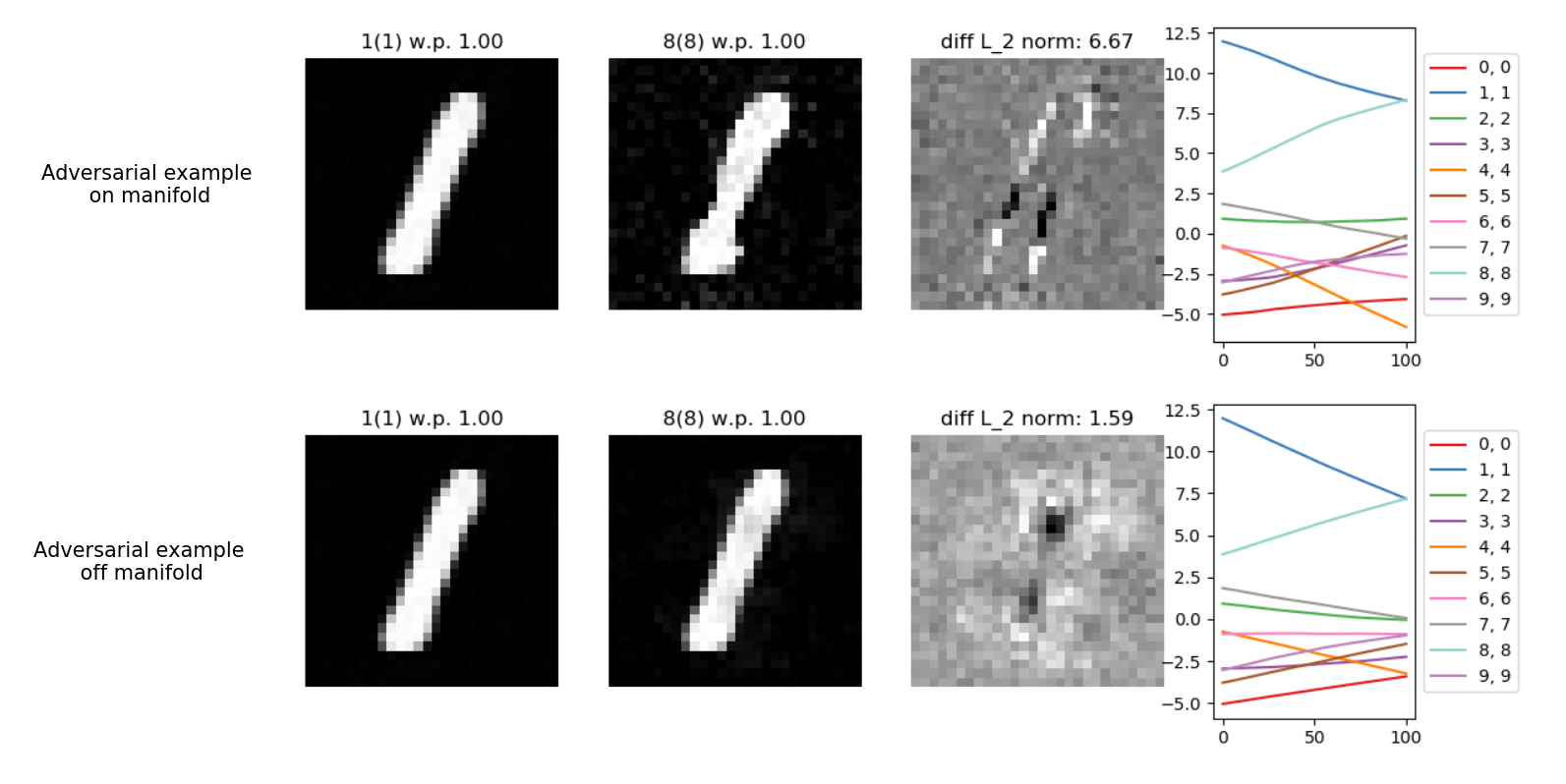}
  \caption{MNIST separate attacks - changing 1 into 8. The on-manifold and the off-manifold attacks are using two different strategies: The on-manifold perturbation tries to control the line's width, whereas the off-manifold perturbation tries to add two black holes at the top and bottom of the 1.}
\end{figure}

\begin{figure}[ht!]
  \centering
  \includegraphics[width=\textwidth]{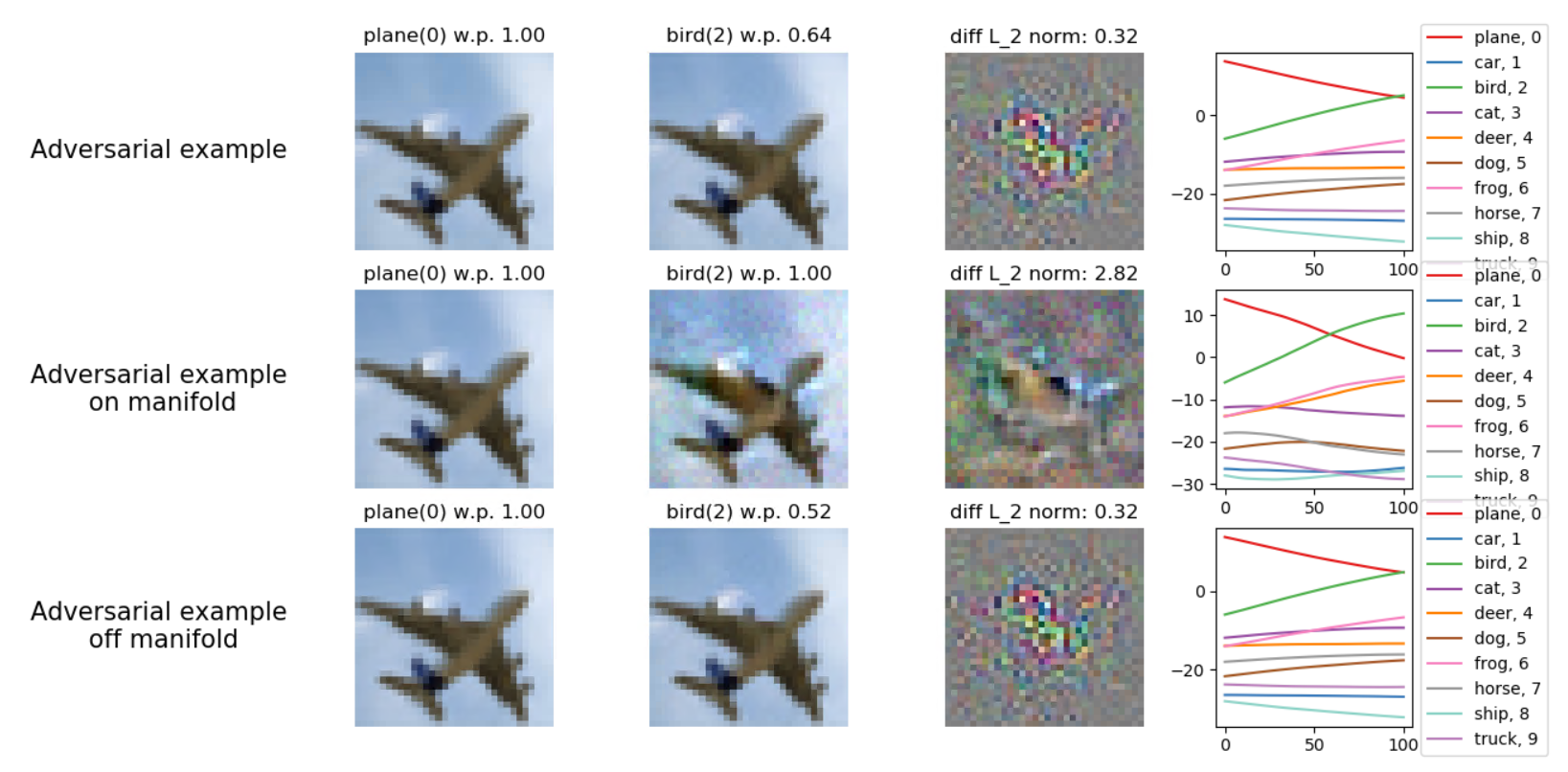}
  \caption{CIFAR separate attacks - changing plane into bird. The on-manifold perturbation is changing the shape of the wings by adding blue to the front part of it. The plane's head is turning into a bird's head by coloring it and creating a narrow neck.}
\end{figure}

\begin{figure}[ht!]
  \centering
  \includegraphics[width=\textwidth]{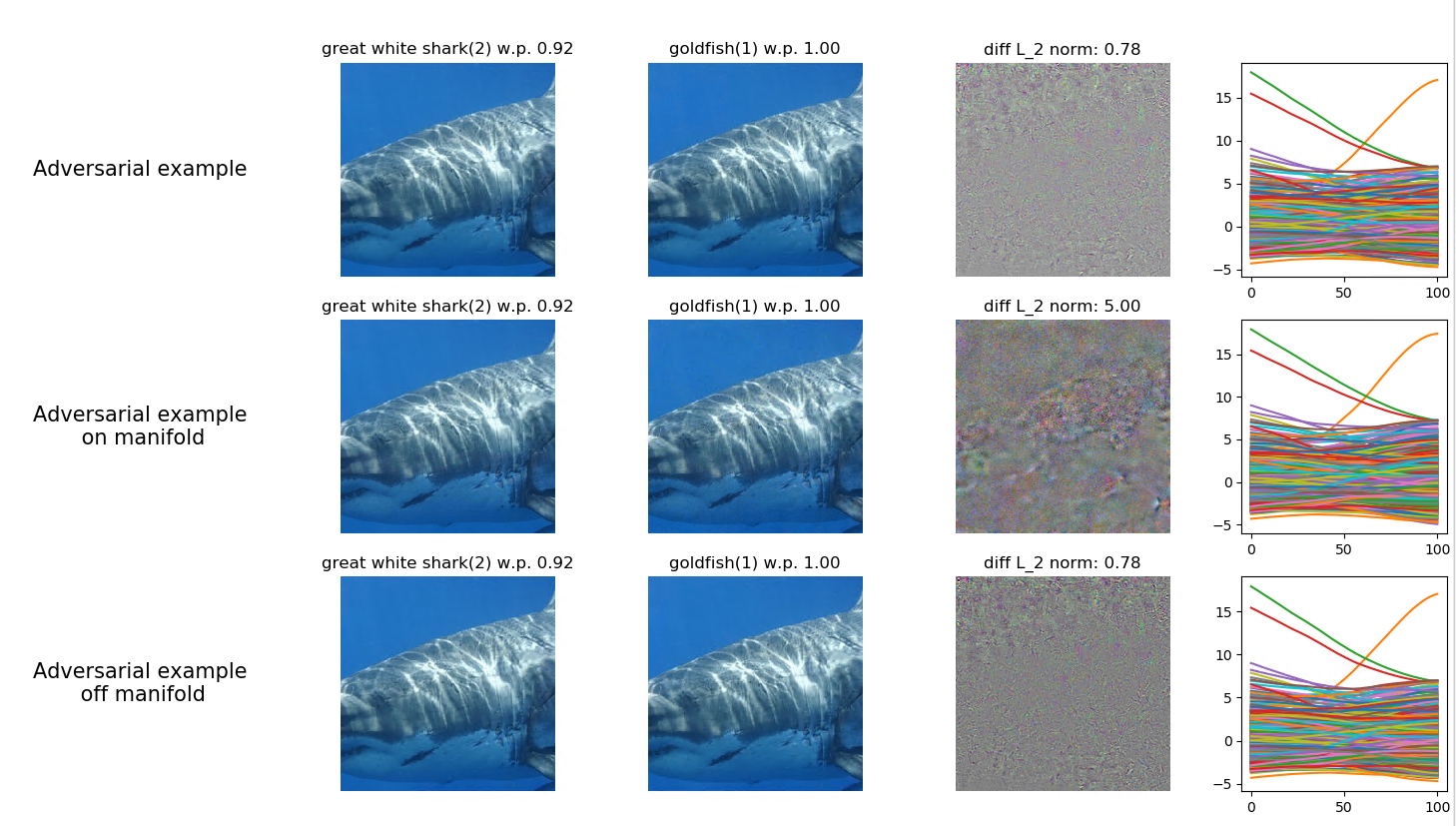}
  \caption{ImageNet separate attacks - changing white shark into goldfish. A smaller fish-shaped noise is added. See further explanation in \ref{subs:imagenet}.}
\end{figure}

\begin{figure}[ht!]
  \centering
  \includegraphics[width=\textwidth]{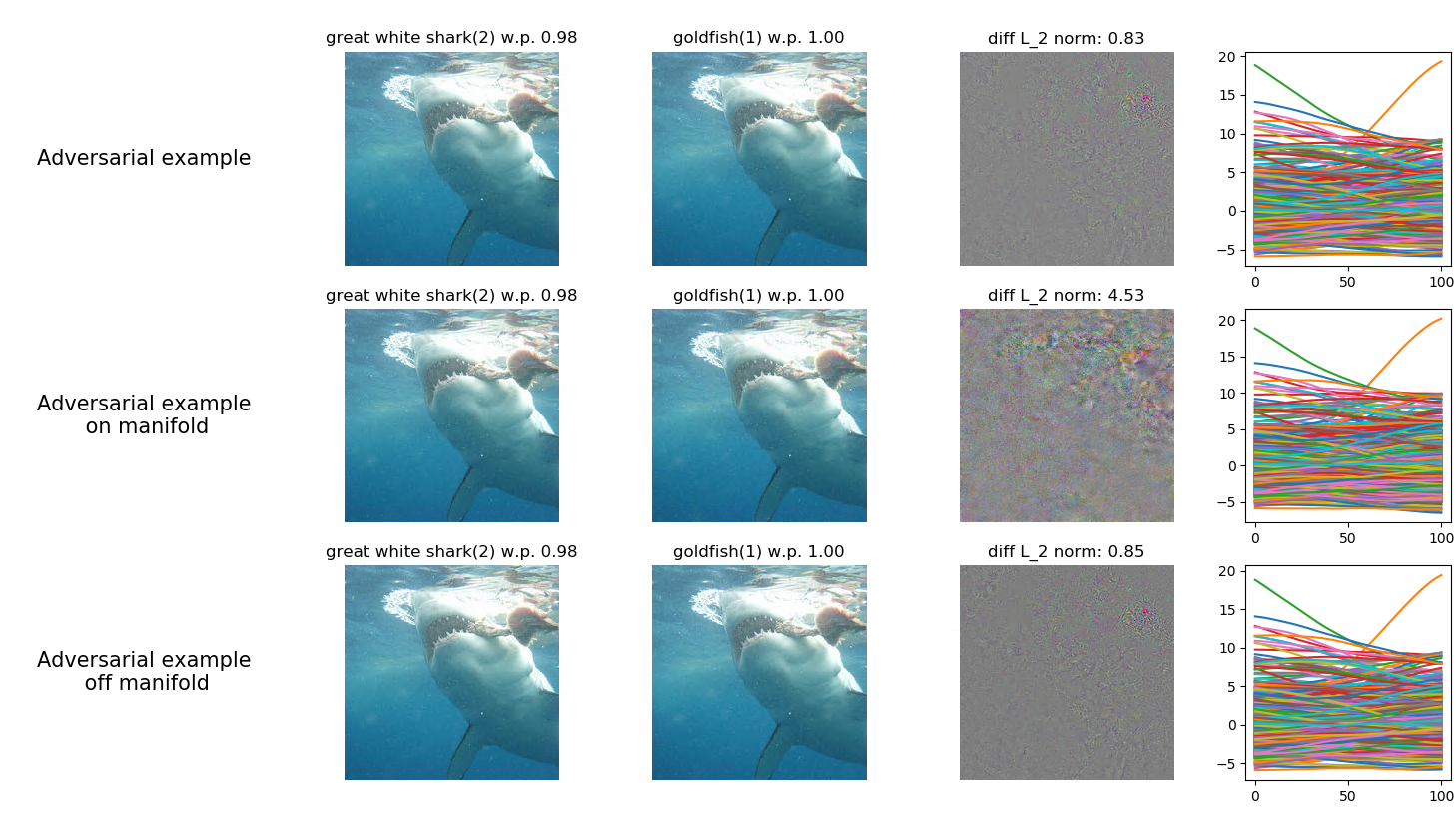}
  \caption{ImageNet separate attacks - changing white shark into goldfish. The small object in the shark's jaws is painted in orange to resemble a goldfish. See further explanation in \ref{subs:imagenet}.}
\end{figure}

\clearpage 

\subsection{ImageNet visual examples}
\label{subs:imagenet}
We have shown visual examples from MNIST CIFAR10 and ImageNet dataset which have similar characteristics when projected on and off the natural image manifold. First, the on-manifold component of the perturbation has a much smaller norm since the perturbation is almost perpendicular to the image manifold. Consequently, it has very little adversarial effect on the classification. Finally, we showed that the adversarial noise on the manifold (in both experiments) is trying to change the image in a visually meaningful way, making it look more like a similarly shaped object from the target class. 

One big difference between the ImageNet dataset and the two others is the image dimensions: The images in ImageNet are of dimension $224 \times 224 \times 3 = 150,528$, which is $192$ and $49$ times bigger than in MNIST and CIFAR10 datasets, respectively. Therefore, in order to change the visual appearance of the image we have to change many more pixels, causing a very large $L_2$ distance. Even though the adversarial perturbation calculated using the on-manifold PGD attack has a much larger norm than the unrestricted  adversarial perturbation, it is still too small to create a visually noticeable difference in such high-dimensional images.


%% file: apdix3.5-lowdimension.tex
\section{Appendix - Low-dimensional experiments}
\label{appendix:low-dim-exp}

\subsection{Model details}
We use the same 2-layer network for all our synthetic experiments. The input is either $[0,1]^2$ or $[0,1]^3$ for the 2-dimensional and 3-dimensional experiments, respectively. The first hidden layer is linear with width 4000 (default initialization), followed by ReLu activation. The second layer is the output layer (4000 input features and 1 output logit), with random fixed weights u.a.r from {-1,1}. The binary classification is received through a simple round of the Sigmoid function of the output logit. 

We trained the network for $200$ epochs, with SGD optimizer. The learning rate is $0.02$ and wight decay is $0.01$.

%% file: apdix4-approx_manifold.tex
\section{Appendix - approximated manifold calculations}
\label{appendix:approx-manifold}

\subsection{Auto-encoders and the local manifold}
In order to talk about on-manifold and off-manifold projections, we have to determine the local orientation of the natural image manifold around some natural image $x$. Though it is commonly believed that such a manifold exists, there is no known way to find its exact representation. However, there are many methods to approximate the natural image manifold, and we decided to use autoencoders to approximate it in the vicinity of each $x$.

Generally speaking, an autoencoder consists of an encoder and a decoder. Given a natural image $x$ the encoder computes a "code" $c$ representing this image in a $k$-dimensional latent space with $k<<n$. The decoder then translates $c$ back to an image $x'$, which is as close as possible to the original image $x$. In this way, the latent space created by the encoder is translated by the decoder into a $k$-dimensional manifold within the image input space, which we denote by $M$.

\subsection{Local linear approximation of the manifold}
For a small epsilon the classifier function in the multi-dimensional epsilon ball around a natural image is almost linear. We use this property when we calculate the projections on and off the manifold. 

Given a test-set image $x$, we first calculate it's code $c$ using the encoder, and the auto-encoded image $x'$ using the decoder. The code $c$ is a much lower dimensional vector, whose dimension is determined by the dimension $k$ of the latent space. In this notation, $x, x' \in [0,1]^n$ and $c \in \mathbb{R}^k$ for $k << n$. 

We want to calculate the linear approximation of the manifold $M$ around $x'$. For each $i \in \{0,...,k-1\}$ and for some small $\epsilon$, we define $\epsilon_i$ to be the k-dimensional vector with $\epsilon$ on the $i$-th entry and zero elsewhere. Next, we look at the code

$$c_i = c + \epsilon_i $$

We decode $c_i$ and denote the resulted image as $x' + s_i$. We next use these $k$ $s_i$'s to span the $k$-dimensional local linear approximation of the manifold around $x'$, denoted by $M = span(s_1,...,s_{k-1})$. As $x$ and $x'$ are optimally close, we use $M$ as an approximation for the local natural image manifold around $x$. When looking for the on-manifold projection of a short vector originated in $x$, we project the vector on the locally linear manifold $M$.

\subsection{Image manifold dimensions}
For \textbf{MNIST} dataset, containing images of $n = 28\times28 = 784$ dimensions we trained an autoencoder with $k = 16$-dimensional latent space (\citet{iae}). For the CIFAR10 and the ImageNet experiments we trained an autoencoder built using the VGG architecture (\citet{VGG}). For simplicity, we trained it only on 2 arbitrarily chosen classes. Therefore, we run these experiments only on correctly classified images from these classes (instead of all test-set images). For \textbf{CIFAR10} (whose images have $n = 3\times 32\times 32 = 3072$ dimensions) we chose the classes "plane" and "car" (indices 0 and 1), encoded into $128$-dimensional latent space. For \textbf{ImageNet} (whose images have $n = 3\times 224\times 224 = 150,528$ dimensions) we trained it on "goldfish" and "white sharks" (indices 1 and 2), resulting in $3584$-dimensional latent space.


%% file: apdix4.5-perpen_exprmnt.tex
\section{Appendix - adversarial examples are roughly perpendicular to the image manifold}
\label{appendix-perpendicularity}
\subsection{Low dimensional experiments}
\begin{figure}
\centering
\begin{subfigure}[b]{0.3\linewidth}
\label{2d_orth}
\centering
\includegraphics[width=\textwidth]{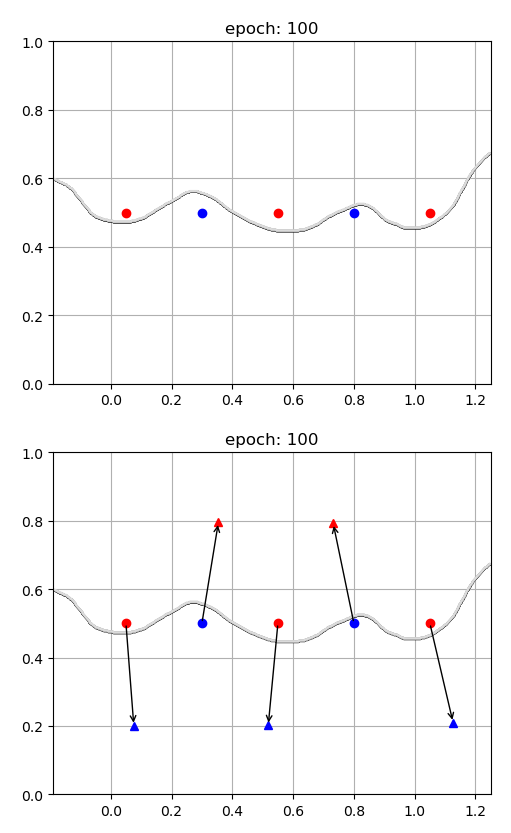}
\caption{2D binary decision boundary, 1D data}
\end{subfigure}

\begin{subfigure}[b]{0.3\linewidth}
\label{3d_orth}
\centering
\includegraphics[width=\textwidth]{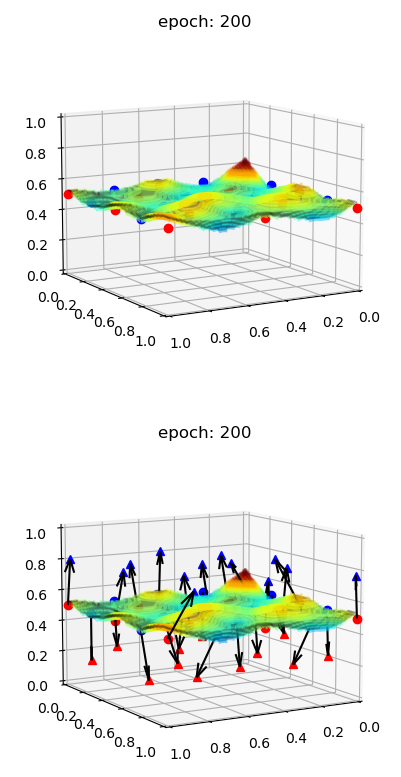}
\caption{3D binary decision boundary, 2D data}
\end{subfigure}
\caption{A binary decision boundary of NNs, adding arrows to represent the gradient direction (but not its real size) from each train input. The actual adversarial perturbations we generate point roughly in these directions but have much smaller norms.}
\end{figure}

In the low dimensional experiments the relevant manifolds are clearly shown in the figures. In the 2D experiment in Figure \ref{2d_orth} the dots are the train data points from two opposite classes marked in red and blue. The grey lines represent the decision boundary and the arrows represent the adversarial direction (i.e. the gradient direction w.r.t. the input calculated at the data point, with normalized norms). One can see that the train data is all within the one-dimensional manifold $y=0.5$. The adversarial direction from each data point is clearly almost perpendicular to the manifold.

In the 3D experiment in Figure \ref{3d_orth} the red and blue dots are the train data points from opposite classes, the colored manifold represent the decision boundary in the $[0,1]^3$ cube, colored by its $z$ values to look like a topographic map. The arrows represent the adversarial direction at each data point. The data is all within the manifold $z=0.5$, and one can clearly see that the arrows are all pointing roughly up or down - namely in primarily off-manifold directions.

\subsection{High dimensional experiments}
In the following sections we experimentally verify the second property (that adversarial directions are almost perpendicular to this manifold) by measuring the off-manifold and on-manifold norms of adversarial perturbations. Given a correctly classified test-set image $x$ and a local manifold $M$ around $x$, we first found an adversarial example $x+d$ (where $\norm{d}_2 \leq \epsilon$ for some small $\epsilon$, and the classifier classifies $x$ and $x+d$ differently). We then projected the adversarial perturbation $d$ onto $M$, and computed the ratio $R(d,M)$ between the norm of $d$ and the norm of its projection $d_{on}$. Note that for a random vector $r= (r_1, ... , r_{n})$ with $r_i \sim N(0,1)$ of dimension $n$, its norm is equally distributed among all the dimensions. Therefore, the expected ratio between $\norm{r}$ and $\norm{r_{on}}$ for a random vector $r$ (which we denote by $\bar{R}$) will be about $\sqrt{\frac{n}{k}}$ (see details in Appendix \ref{appendix:random-vector}). For $k<<n$ this expected ratio is large, making most vectors almost perpendicular to small manifolds by definition. 

In our experiments we measured the average of the ratio $R(d,M)$ where $d$ is an adversarial perturbation and $M$ is the natural image manifold, and compared it to $\bar{R}$ (note that values larger than $\bar{R}$ indicate actual preference of  off-manifold over on-manifold dimensions by the DNN, but any ratio larger than $\sqrt{2}$ indicates an angle larger than 45 degrees). In this paper we use the Euclidean norm in all our attacks; experiments with $L_{\infty}$-norm adversarial perturbations can be found in appendix \ref{appendix:linfinity}.

    For an image $x$, we started with calculating an adversarial example $x+d$ using a multi-step PGD attack (\citet{PGD}). We projected $d$ onto the locally approximated linear manifold $M$, measured the aforementioned average ratio $R(d, M) = \frac{\norm{d}}{\norm{Proj_{M}(d)}}$, and compared it to the ratio $\bar{R}$ which is expected for random vectors.
    
\paragraph{The relative effect of the projected vector on the change in classification}

To measure the effect of the on-manifold projection on the change of classification for the classifier
function $f$ (which is the neural network without the SoftMax final layer), an original class $i$, and
an adversarial class $t$, we first define the total effect of the adversarial vector $d$ as

$$
E_d = f(x'+d)_t - f(x'+d)_i - (f(x')_t - f(x')_i)
$$
    
Finally, we calculate the on-manifold relative effect out of the total adversarial effect as
    
$$
E_{Proj_M(d)} = \frac{f(x'+Proj_M(d))_t - f(x'+Proj_M(d))_i - (f(x')_t  - f(x')_i)}{E_d}
$$
    
\begin{table}[ht!]
  \caption{The norm ratio and on-manifold relative effect for natural images}
  \label{tab:natural-norm-ratio}
  \centering
  \begin{tabular}{llll}
    \toprule
    Dataset     &  $\bar{R}$     &  $R(d, M)$ & $E_{Proj_M(d)}$ \\
    \midrule
    MNIST & $\sqrt{\frac{n}{k}} = 7$  & 3.68  &  8\% \\
    CIFAR10     & $\sqrt{\frac{n}{k}} = 4.89$ & 5.59   & 4\%   \\
    ImageNet     & $\sqrt{\frac{n}{k}} = 6.48$       & 5.5 & 2\% \\
    \bottomrule
  \end{tabular}
\end{table}

In Table \ref{tab:natural-norm-ratio} one can see that for MNIST dataset the norm ratio was smaller than expected. The reason for it may be that the MNIST dataset can be successfully classified using a linear classifier, making it similar to the linear synthetic case in which the ratio was 1. Note that in all three cases, the on-manifold relative effect was tiny, demonstrating that the networks primarily used the off-manifold perturbations rather than the on-manifold perturbations in order to switch their classification.

 \subsection{\textbf{Experiment 2: Adversarial examples with on/off manifold constraints}}
    In this experiment, starting from an image $x$, we used the PGD attack with an extra constraint - before normalizing each step onto the epsilon-step sphere, we projected it on or off the approximated local linear manifold $M$ (see Appendix \ref{appendix:onoffattacks} for attack details). In this way, we generated on and off manifold adversarial examples, and compared the resultant adversarial distances between the three attacks: unconstrained PGD, on-manifold PGD and off-manifold PGD.

\begin{table}[ht!]
  \caption{The mean distance of an adversarial example}
  \label{tab:mean-distance-adv}
  \centering
  \begin{tabular}{llll}
    \toprule
    Dataset &  no-constraint PGD &  On-manifold PGD &  Off-manifold PGD \\
    \midrule
    MNIST & 1.01  & \textbf{3.68 \quad (364\%)}  &  1.05 \quad (103\%) \\
    CIFAR10     & 0.3 & \textbf{1.87 \quad (623\%)}   & 0.305 \quad (101\%)   \\
    ImageNet     & 0.149 & \textbf{0.849 \quad (570\%)} & 0.152 \quad (102\%)\\
    \bottomrule
  \end{tabular}
\end{table}

    In Table \ref{tab:mean-distance-adv} one can see a big difference in the adversarial distances. In all datasets, on-manifold adversarial examples are much further away from the original image $x$ while the off-manifold adversarial examples are much closer (up to 6 times closer).

%% file: apdix5-onoffPGD.tex
\section{Appendix - separate on and off manifold PGD attack details}
\label{appendix:onoffattacks}
The multi-step PGD attack \citet{PGD} as implemented in the \textit{Advertorch} python package, consists of 4 main parts in each step:

\begin{enumerate}
    \item Calculate the network's gradient with respect to the input image $x$.
    \item Normalize the gradient to be $\epsilon$-step sized vector with respect to the chosen norm.
    \item Add the current calculated step to the accumulative adversarial vector.
    \item Clip the total adversarial vector to be at most $\epsilon$ sized vector (and each pixel to be $\in [0,1]$).
\end{enumerate}

For a multi-step PGD attack on natural images we used the $L_2$ norm, up to 50 steps, the margin loss function and epsilon of 2, 0.5, and 2 for MNIST, CIFAR10, and ImageNet respectively. The step size is $0.04$ for MNIST and CIFAR10 and $0.8$ for ImageNet. We run a greedy version of the attack - we stop as soon as we get to an adversarial example (change in classification).

In order to get separate on and off manifold adversarial examples, we project each gradient step respectively on or off the manifold before normalization. Therefore, the new step is:

\begin{enumerate}
    \item Calculate the network's gradient with respect to the input image $x$.
    \item \textbf{Project the gradient step vector on or off the manifold.}
    \item Normalize the gradient to be $\epsilon$-step sized vector with respect to the chosen norm.
    \item Add the current calculated step to the accumulative adversarial vector.
    \item Clip the total adversarial vector to be at most $\epsilon$ sized vector (and each pixel to be $\in [0,1]$).

\end{enumerate}

For the multi-step on or off PGD attack on natural images we used the $L_2$ norm, up to 1000 steps, the margin loss function and epsilon of 10, 3, and 5 for MNIST, CIFAR10, and ImageNet respectively. The step size is $0.02$, $0.12$ and $0.01$ for MNIST, CIFAR10, and ImageNet respectively for the regular and the off-manifold attacks. The step size is multiplied by $10$ for the on-manifold attack, in order to reach further. Here also, we run a greedy version of the attack.


%% file: apdix6-NNdetails.tex
\section{Appendix - image auto-encoders and image classifiers details}
\label{appendix:architecture}
\subsection{MNIST dataset}
We used two hidden layers classification network:
\begin{enumerate}
    \item Linear layer (in=784, out=256), ReLU layer, Dropout layer w.p. 0.2 (drop out entries with 0.2 probability)
    \item Linear layer (in=256, out=256), ReLU layer, Dropout layer w.p. 0.2 (drop out entries with 0.2 probability)
    \item Linear output layer (in=256, out=10)
\end{enumerate}

The NN was trained using SGD optimizer with learning rate of $0.01$, weight decay of $0.0001$, batch size of $200$ and for 40 epochs, reaching 98\% success in classification.

The MNIST encoder and decoder are described in \citet{iae}. They were trained using Adam optimizer for 1000 epochs, with learning rate of $0.001$ and batch size of $144$. The chosen latent space dimension is 16. The result mean MSE loss for the $10,000$ test-set images is $0.00003$, and the average $L_2$ distance between the MNIST images and its matching autoencoded images is $2.1$.\\

\subsection{CIFAR10 dataset}
For the CIFAR10 experiments we used a pre-trained 7 layered convolutional classifier network (which can be found at https://github.com/aaron-xichen/pytorch-playground) achieving 93\% success rate in its classification. 

For the CIFAR10 autoencoder we used an autoencoder created for cars dataset from Stanford by Yang Liu (found here https://github.com/foamliu/Autoencoder) and based on the VGG16 architecture. In order to get a low-dimensional latent space, we replaced the max-up-pooling decoder layers (that leak additional information from the encoder side to the decoder side in the form of max-pool indices, thus increasing the effective width of the bottleneck) with regular up-sampling layers with the same sample ratio. We trained it only for the classes of plains (0) and cars (1) using Adam optimizer for 400 epochs, with learning rate of $0.001$, batch size of $32$ and a latent space of $128$ dimensions. The result mean MSE loss for the $2000$ test-set images is $0.0001$, and the average $L_2$ distance between these images and their matching autoencoded images is $7.88$.\\

\subsection{ImageNet dataset}
For the ImageNet experiments we used a pre-trained ResNet50 network from the \textit{pytorch} package (\citet{resnet}).

For the ImageNet autoencoder we used the same architecture as in CIFAR10. It was trained using Adam optimizer for 325 epochs, with learning rate of $0.0001$ and batch size of $32$. The auto encoder was trained to use a latent space of dimension $3584$. The result mean MSE loss for the $100$ test-set images is $0.0001$, and the average $L_2$ distance between these images and its matching autoencoded images is $38.7601$.\\

%% file: apdix8-inf_norm.tex
\section{Appendix - comparison to $L_\infty$ norm}
\label{appendix:linfinity}
\begin{figure}[htp!]
    \centering
    \includegraphics[width=8cm]{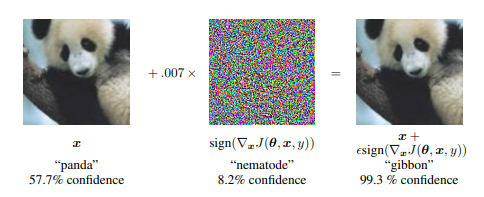}
    \caption{\citet{goodfellow2014explaining} $L_\infty$ adversarial example}
    \label{fig:panda}
\end{figure}

This paper discusses adversarial attacks using the Euclidean norm, while most of the literature describes and discusses $L_\infty$ norm attacks. The common depiction of an adversarial example consists of some natural image to which we add a tiny randomly-looking perturbation. A typical example is the panda bear (from \citet{goodfellow2014explaining}), where the adversarial perturbation added to the image looks like random noise which is multiplied by a very small constant factor (Figure \ref{fig:panda}).

\begin{figure}[htp!]
    \centering
    \includegraphics[width=8cm]{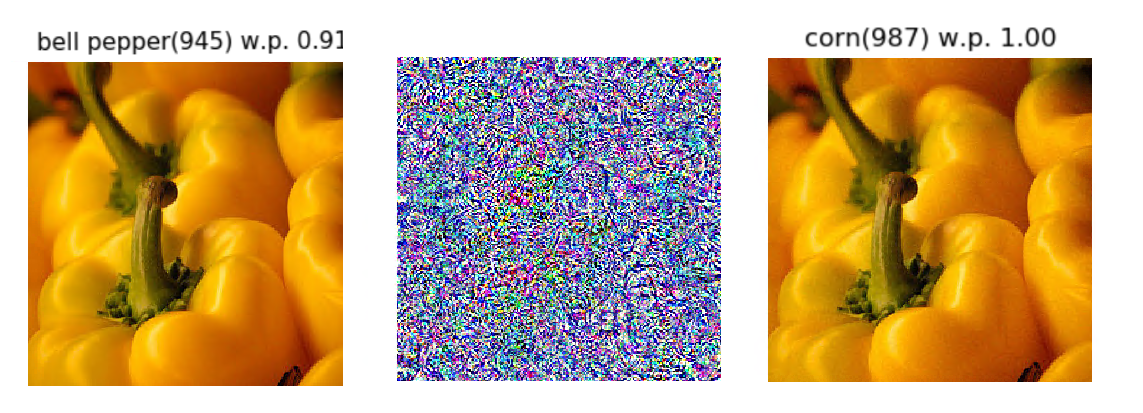}
    \caption{multi-step PGD $L_\infty$ adversarial example}
    \label{fig:linfexp}
\end{figure}

Running the multi-step PGD attack (\citet{PGD}) on an arbitrary test-set image from the ImageNet dataset also generates a similar randomly-looking perturbation (Figure \ref{fig:linfexp}).

\begin{figure}[htp!]
    \centering
    \includegraphics[width=8cm]{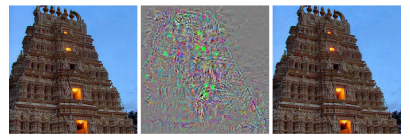}
    \caption{\citet{szegedy2013intriguing} $L_2$ adversarial example}
    \label{fig:ostrich}
\end{figure}

When discussing an $L_2$ adversarial attack, the adversarial perturbation seems much less random. In the paper of \citet{szegedy2013intriguing} there are adversarial examples created by an $L_2$ attack, presenting a perturbation that fits the object's general outline and expands in a halo around it (Figure \ref{fig:ostrich}).

\begin{figure}[htp!]
    \centering
    \includegraphics[width=8cm]{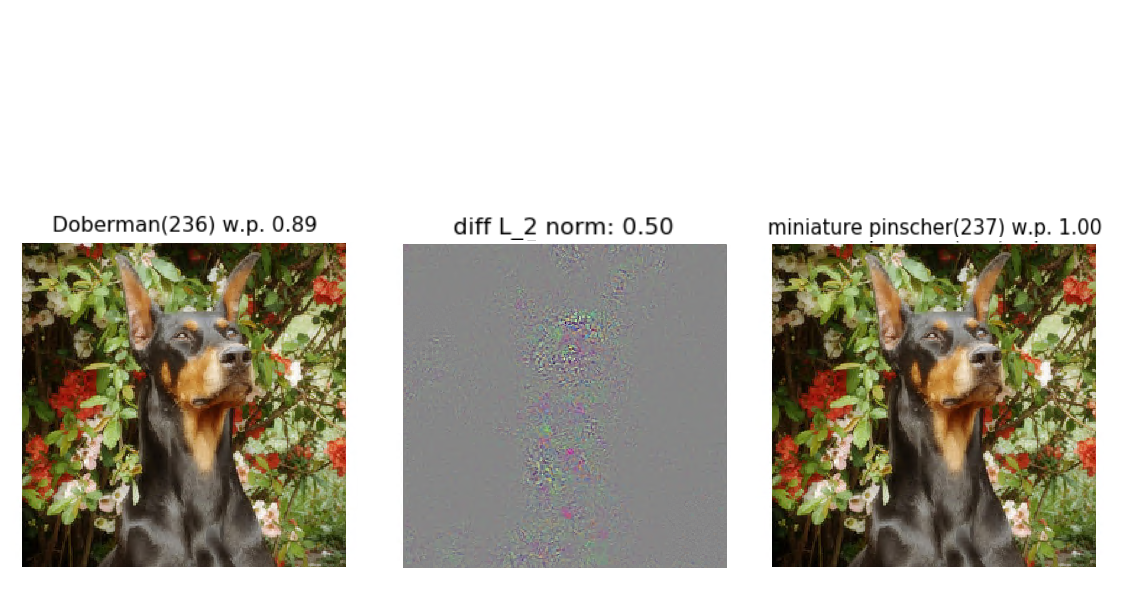}
    \caption{\citet{szegedy2013intriguing} $L_2$ adversarial example}
    \label{fig:l2exp}
\end{figure}

Running the original PGD attack on an arbitrary test-set image from the ImageNet dataset demonstrate the same object-oriented perturbation (Figure \ref{fig:l2exp})

Why does the $L_\infty$ adversarial perturbation look like random noise? The answer is related to the $L_\infty$ normalization of the gradient-step. When using FGSM attack (\citet{goodfellow2014explaining}) or PGD attack (\citet{PGD}) with perturbation size $\epsilon$ ($L_\infty$-norm), we first calculate the best adversarial direction using gradient - that is the local optimal noise we should add to the natural image in order to change its classification. The gradients in both attacks are calculated for each pixel of the natural image. When performing $L_\infty$ normalization, we take any pixel change - tiny or large - to be $\epsilon$-sized. Note, the $L_\infty$-normalized gradient is no longer the optimal direction, as normalization changes the vector's original direction.

The $L_2$-normalizations performed in the $L_2$ versions of the attacks, of course, don't change the gradient direction. Therefore, we can understand that the optimal perturbation applies more significant changes to pixels that play more prominent roles in the classification. Consequently, most of the changes happen within the figure, especially around its edges, which makes the perturbation have the general appearance of the original image.

$L_\infty$ attacks behave differently. Due to $L_\infty$ normalization, the background pixel's tiny noise and the object-surrounding pixel's larger noise are both becoming $\epsilon$-sized noise. The $L_\infty$ perturbations thus look like random noise with almost uniform amplitude, which appears both inside and outside the classified object.

%% file: apdix2-random_projection.tex
\section{Appendix - random-vector projection ratio}
\label{appendix:random-vector}
We look at a random vector $r= (r_1, ... , r_{n})$ for $r_i \sim N(0,1)$ with dimension $n$. For convenience, we normalize $r$ by dividing it by its $L_2$ norm. 

Due to symmetry, We can rotate $r$ so that its first $k$ entries represent it's on-manifold projection  - 
$$ r_{on} = Proj_{N}(r) = (r_1, ... , r_{k}, 0, ..., 0)$$

and its next $n-k$ entries represent its off-manifold projection - 
$$ r_{off} = Proj_{N^\perp}(r) = r - (r_1, ... , r_{k}, 0, ..., 0) = (0, ... , 0, r_{k+1}, ..., r_n)$$

Note, $1= \norm{r}_2 = \sqrt{r_1^2 + ... + r_{n}^2}$, $\norm{r_{on}}_2 = \sqrt{r_1^2 + ... + r_{k}^2}$, $\norm{r_{off}}_2 = \sqrt{r_{k+1}^2 + ... + r_{n}^2}$. When looking at the ratio between $r$ and its projection on the manifold $r_{on}$ we get - 

$$\frac{\norm{r}}{\norm{r_{on}}} = \sqrt{\frac{r_1^2 + ... + r_{n}^2}{r_1^2 + ... + r_{k}^2}} =\sqrt{ 1 + \frac{r_{k+1}^2 + ... + r_{n}^2}{r_1^2 + ... + r_{k}^2}} $$

All $r_i$s are independent and have the same distribution, so all $r_i^2$s are also i.i.d. Therefore, if we look at the expectation of the squared ratio we get -

$$\mathbb{E}[\frac{\norm{r}^2}{\norm{r_{on}}^2}] = \mathbb{E}[1 + \frac{r_{k+1}^2 + ... + r_{n}^2}{r_1^2 + ... + r_{k}^2}] = 1 + \frac{\mathbb{E}[r_{k+1}^2] + ... + \mathbb{E}[r_{n}^2]}{\mathbb{E}[r_1^2] + ... + \mathbb{E}[r_{k}^2]} =$$
$$ 1 + \frac{(n-k)\cdot \mathbb{E}[r_{i}^2]}{k \cdot \mathbb{E}[r_j^2]} = 1 + \frac{n-k}{k} = \frac{n}{k} $$

In conclusion, we can upper bound the expectation of the ratio between the norms - 
$$\mathbb{E}[\frac{\norm{r}}{\norm{r_{on}}}] \leq \sqrt{\frac{n}{k}} $$

\newpage

%% file: main.bbl
\begin{thebibliography}{28}
\providecommand{\natexlab}[1]{#1}
\providecommand{\url}[1]{\texttt{#1}}
\expandafter\ifx\csname urlstyle\endcsname\relax
  \providecommand{\doi}[1]{doi: #1}\else
  \providecommand{\doi}{doi: \begingroup \urlstyle{rm}\Url}\fi

\bibitem[Atzmon et~al.(2020)Atzmon, Gropp, and Lipman]{iae}
M.~Atzmon, A.~Gropp, and Y.~Lipman.
\newblock Isometric autoencoders.
\newblock \emph{arXiv preprint arXiv:2006.09289}, 2020.

\bibitem[Biggio et~al.(2013)Biggio, Corona, Maiorca, Nelson, {\v{S}}rndi{\'c},
  Laskov, Giacinto, and Roli]{biggio2013evasion}
B.~Biggio, I.~Corona, D.~Maiorca, B.~Nelson, N.~{\v{S}}rndi{\'c}, P.~Laskov,
  G.~Giacinto, and F.~Roli.
\newblock Evasion attacks against machine learning at test time.
\newblock In \emph{Joint European conference on machine learning and knowledge
  discovery in databases}, pages 387--402. Springer, 2013.

\bibitem[Bourlard and Kamp(1988)]{bourlard1988auto}
H.~Bourlard and Y.~Kamp.
\newblock Auto-association by multilayer perceptrons and singular value
  decomposition.
\newblock \emph{Biological cybernetics}, 59\penalty0 (4):\penalty0 291--294,
  1988.

\bibitem[Fawzi et~al.(2016)Fawzi, Moosavi-Dezfooli, and
  Frossard]{fawzi2016robustness}
A.~Fawzi, S.-M. Moosavi-Dezfooli, and P.~Frossard.
\newblock Robustness of classifiers: from adversarial to random noise.
\newblock \emph{Advances in Neural Information Processing Systems}, 29, 2016.

\bibitem[Gilmer et~al.(2018)Gilmer, Metz, Faghri, Schoenholz, Raghu,
  Wattenberg, and Goodfellow]{gilmer2018adversarial}
J.~Gilmer, L.~Metz, F.~Faghri, S.~S. Schoenholz, M.~Raghu, M.~Wattenberg, and
  I.~Goodfellow.
\newblock Adversarial spheres.
\newblock \emph{arXiv preprint arXiv:1801.02774}, 2018.

\bibitem[Goodfellow(2017)]{goodfellowtalk}
I.~Goodfellow.
\newblock Stanford university school of engineering - lecture 16 | adversarial
  examples and adversarial training, 2017.
\newblock URL \url{https://www.youtube.com/watch?v=CIfsB_EYsVI&t=3s}.

\bibitem[Goodfellow et~al.(2014)Goodfellow, Shlens, and
  Szegedy]{goodfellow2014explaining}
I.~J. Goodfellow, J.~Shlens, and C.~Szegedy.
\newblock Explaining and harnessing adversarial examples.
\newblock \emph{arXiv preprint arXiv:1412.6572}, 2014.

\bibitem[He et~al.(2016)He, Zhang, Ren, and Sun]{resnet}
K.~He, X.~Zhang, S.~Ren, and J.~Sun.
\newblock Deep residual learning for image recognition.
\newblock In \emph{Proceedings of the IEEE conference on computer vision and
  pattern recognition}, pages 770--778, 2016.

\bibitem[Ilyas et~al.(2019{\natexlab{a}})Ilyas, Santurkar, Tsipras, Engstrom,
  Tran, and Madry]{ilyas2019adversarial}
A.~Ilyas, S.~Santurkar, D.~Tsipras, L.~Engstrom, B.~Tran, and A.~Madry.
\newblock Adversarial examples are not bugs, they are features.
\newblock \emph{Advances in neural information processing systems}, 32,
  2019{\natexlab{a}}.

\bibitem[Ilyas et~al.(2019{\natexlab{b}})Ilyas, Santurkar, Tsipras, Engstrom,
  Tran, and Madry]{madrysexpr}
A.~Ilyas, S.~Santurkar, D.~Tsipras, L.~Engstrom, B.~Tran, and A.~Madry.
\newblock Adversarial examples are not bugs, they are features.
\newblock \emph{arXiv preprint arXiv:1905.02175}, 2019{\natexlab{b}}.

\bibitem[Jalal et~al.(2017)Jalal, Ilyas, Daskalakis, and
  Dimakis]{jalal2017robust}
A.~Jalal, A.~Ilyas, C.~Daskalakis, and A.~G. Dimakis.
\newblock The robust manifold defense: Adversarial training using generative
  models.
\newblock \emph{arXiv preprint arXiv:1712.09196}, 2017.

\bibitem[Khoury and Hadfield-Menell(2018)]{khoury2018geometry}
M.~Khoury and D.~Hadfield-Menell.
\newblock On the geometry of adversarial examples.
\newblock \emph{arXiv preprint arXiv:1811.00525}, 2018.

\bibitem[Krizhevsky(2009)]{CIFAR10}
A.~Krizhevsky.
\newblock Learning multiple layers of features from tiny images.
\newblock Technical report, 2009.

\bibitem[LeCun et~al.(2010)LeCun, Cortes, and Burges]{lecun2010mnist}
Y.~LeCun, C.~Cortes, and C.~Burges.
\newblock Mnist handwritten digit database.
\newblock \emph{ATT Labs [Online]. Available:
  http://yann.lecun.com/exdb/mnist}, 2, 2010.

\bibitem[Madry et~al.(2018)Madry, Makelov, Schmidt, Tsipras, and Vladu]{PGD}
A.~Madry, A.~Makelov, L.~Schmidt, D.~Tsipras, and A.~Vladu.
\newblock Towards deep learning models resistant to adversarial attacks.
\newblock \emph{ICLR}, 2018.

\bibitem[Mahloujifar et~al.(2019)Mahloujifar, Diochnos, and
  Mahmoody]{mahloujifar2019curse}
S.~Mahloujifar, D.~I. Diochnos, and M.~Mahmoody.
\newblock The curse of concentration in robust learning: Evasion and poisoning
  attacks from concentration of measure.
\newblock In \emph{Proceedings of the AAAI Conference on Artificial
  Intelligence}, volume~33, pages 4536--4543, 2019.

\bibitem[Meng and Chen(2017)]{meng2017magnet}
D.~Meng and H.~Chen.
\newblock Magnet: a two-pronged defense against adversarial examples.
\newblock 2017.

\bibitem[Pope et~al.(2021)Pope, Zhu, Abdelkader, Goldblum, and
  Goldstein]{pope2021intrinsic}
P.~Pope, C.~Zhu, A.~Abdelkader, M.~Goldblum, and T.~Goldstein.
\newblock The intrinsic dimension of images and its impact on learning.
\newblock \emph{ICLR 2021}, 2021.

\bibitem[Ruderman(1994)]{statnatural}
D.~L. Ruderman.
\newblock The statistics of natural images.
\newblock \emph{Network: Computation in Neural Systems}, 5\penalty0
  (4):\penalty0 517--548, 1994.
\newblock \doi{10.1088/0954-898X\_5\_4\_006}.
\newblock URL \url{https://doi.org/10.1088/0954-898X_5_4_006}.

\bibitem[Russakovsky et~al.(2015)Russakovsky, Deng, Su, Krause, Satheesh, Ma,
  Huang, Karpathy, Khosla, Bernstein, Berg, and Fei-Fei]{IMAGENET}
O.~Russakovsky, J.~Deng, H.~Su, J.~Krause, S.~Satheesh, S.~Ma, Z.~Huang,
  A.~Karpathy, A.~Khosla, M.~Bernstein, A.~C. Berg, and L.~Fei-Fei.
\newblock {ImageNet Large Scale Visual Recognition Challenge}.
\newblock \emph{International Journal of Computer Vision (IJCV)}, 115\penalty0
  (3):\penalty0 211--252, 2015.
\newblock \doi{10.1007/s11263-015-0816-y}.

\bibitem[Samangouei et~al.(2018)Samangouei, Kabkab, and
  Chellappa]{samangouei2018defense}
P.~Samangouei, M.~Kabkab, and R.~Chellappa.
\newblock Defense-gan: Protecting classifiers against adversarial attacks using
  generative models.
\newblock \emph{arXiv preprint arXiv:1805.06605}, 2018.

\bibitem[Shwartz-Ziv and Tishby(2017)]{tishbi}
R.~Shwartz-Ziv and N.~Tishby.
\newblock Opening the black box of deep neural networks via information.
\newblock \emph{arXiv preprint arXiv:1703.00810}, 2017.

\bibitem[Simonyan and Zisserman(2014)]{VGG}
K.~Simonyan and A.~Zisserman.
\newblock Very deep convolutional networks for large-scale image recognition.
\newblock \emph{arXiv preprint arXiv:1409.1556}, 2014.

\bibitem[Stutz et~al.(2019)Stutz, Hein, and Schiele]{stutz2019disentangling}
D.~Stutz, M.~Hein, and B.~Schiele.
\newblock Disentangling adversarial robustness and generalization.
\newblock In \emph{Proceedings of the IEEE/CVF Conference on Computer Vision
  and Pattern Recognition}, pages 6976--6987, 2019.

\bibitem[Szegedy et~al.(2013)Szegedy, Zaremba, Sutskever, Bruna, Erhan,
  Goodfellow, and Fergus]{szegedy2013intriguing}
C.~Szegedy, W.~Zaremba, I.~Sutskever, J.~Bruna, D.~Erhan, I.~Goodfellow, and
  R.~Fergus.
\newblock Intriguing properties of neural networks.
\newblock \emph{arXiv preprint arXiv:1312.6199}, 2013.

\bibitem[Tanay and Griffin(2016)]{tanay2016boundary}
T.~Tanay and L.~Griffin.
\newblock A boundary tilting persepective on the phenomenon of adversarial
  examples.
\newblock \emph{arXiv preprint arXiv:1608.07690}, 2016.

\bibitem[Tsipras et~al.(2018)Tsipras, Santurkar, Engstrom, Turner, and
  Madry]{tsipras2018robustness}
D.~Tsipras, S.~Santurkar, L.~Engstrom, A.~Turner, and A.~Madry.
\newblock Robustness may be at odds with accuracy.
\newblock \emph{arXiv preprint arXiv:1805.12152}, 2018.

\bibitem[Wang et~al.(2014)Wang, Huang, Wang, and Wang]{multilayerAE}
W.~Wang, Y.~Huang, Y.~Wang, and L.~Wang.
\newblock Generalized autoencoder: A neural network framework for
  dimensionality reduction.
\newblock In \emph{2014 IEEE Conference on Computer Vision and Pattern
  Recognition Workshops}, pages 496--503, 2014.
\newblock \doi{10.1109/CVPRW.2014.79}.

\end{thebibliography}
